\documentclass{article}   

\usepackage{microtype}
\usepackage{graphicx}
\usepackage{subfigure}
\usepackage{booktabs} 

\usepackage{hyperref}

\usepackage[accepted]{icml2024}

\usepackage{amsmath}
\usepackage{amssymb}
\usepackage{mathtools}
\usepackage{amsthm}

\usepackage[capitalize,noabbrev]{cleveref}

\theoremstyle{plain}
\newtheorem{theorem}{Theorem}
\newtheorem{proposition}[theorem]{Proposition}
\newtheorem{lemma}[theorem]{Lemma}

\theoremstyle{definition}
\newtheorem{definition}[theorem]{Definition}

\newtheorem{remark}{Remark}

\usepackage[textsize=tiny]{todonotes}

\usepackage{bm}
\usepackage{bbm}  

\usepackage{mathrsfs}
\usepackage{multirow}

\usepackage{comment}

\usepackage{xcolor}
\usepackage{url}
\usepackage{enumitem}

\renewcommand{\hat}{\widehat}
\renewcommand{\epsilon}{\varepsilon}
\def\e{\mathrm{e}}
\def\E{\mathbb{E}}

\def\R{\mathbb{R}}
\def\Rp{\mathbb{R}_{+}} 
\def\N{\mathbb{N}}

\def\calC{\mathcal{C}}

\def\calE{\mathcal{E}}

\def\calG{\mathcal{G}}
\def\calH{\mathcal{H}}

\def\calL{\mathcal{L}}

\def\calN{\mathcal{N}}
\def\calR{\mathcal{R}}
\def\calP{\mathcal{P}}

\def\calS{\mathcal{S}}

\DeclareMathOperator*{\argmin}{arg\,min}

\DeclareMathOperator{\dom}{dom}
\DeclareMathOperator{\interior}{int}

\DeclarePairedDelimiter{\abs}{\lvert}{\rvert} %
\DeclarePairedDelimiter{\brk}{[}{]}
\DeclarePairedDelimiter{\crl}{\{}{\}}
\DeclarePairedDelimiter{\set}{\{}{\}}
\DeclarePairedDelimiter{\prn}{(}{)}
\DeclarePairedDelimiter{\nrm}{\|}{\|}
\DeclarePairedDelimiter{\tri}{\langle}{\rangle}

\newcommand{\innerprod}[2]{\left\langle{#1},{#2}\right\rangle}

\newcommand{\ind}[1]{\mathbbm{1}{\left[#1\right]}}  
\newcommand{\Expect}[1]{\mathbb{E}{\left[#1\right]}}
\newcommand{\Et}[1]{\mathbb{E}_{t}{\left[#1\right]}}

\renewcommand{\d}{\mathrm{d}}

\newcommand{\ones}{\mathbf{1}}

\newcommand{\eg}{\textit{e.g.,}}

\newcommand{\Reg}{\mathsf{Reg}}
\newcommand{\Deltamin}{\Delta}

\newcommand{\lossmat}{\calL}
\newcommand{\fbmat}{\Phi}

\newcommand{\OPT}{\mathsf{OPT}}

\newcommand{\stabtrans}{\mathsf{ST}}  

\newcommand{\psiNS}{\psi^{\mathsf{NS}}}  
\newcommand{\psiLB}{\psi^{\mathsf{LB}}}  
\newcommand{\psiCNS}{\psi^{\mathsf{CNS}}}  
\newcommand{\psiCLB}{\psi^{\mathsf{CLB}}}  

\newcommand{\phiNS}{\phi^{\mathsf{NS}}}  
\newcommand{\phiLB}{\phi^{\mathsf{LB}}}  
\newcommand{\phiCNS}{\phi^{\mathsf{CNS}}}  
\newcommand{\phiCLB}{\phi^{\mathsf{CLB}}}  

\newcommand{\objexo}{\Omega} 
\newcommand{\stab}{\calS} 
 
\newcommand{\yhat}{\hat{y}}

\newcommand{\nn}{\nonumber\\}
\newcommand{\n}{\nonumber}
\newcommand{\per}{\,.}
\newcommand{\com}{\,,}

\newcommand{\sumT}{\sum_{t=1}^T}

\newcommand{\sumak}{\sum_{a=1}^k}

\newcommand{\sumk}{\sum_{i=1}^k}

\newcommand{\prx}[1]{\left(#1\right)}

\newcommand{\brx}[1]{\left\{#1\right\}}
\newcommand{\abx}[1]{\left|#1\right|}

\newcommand{\at}{A_t}
\newcommand{\pt}{p_t}
\newcommand{\pta}{p_{ta}}
\newcommand{\ptAt}{p_{t\at}}

\newcommand{\qt}{q_t}
 
\newcommand{\qta}{q_{ta}}

\icmltitlerunning{Logarithmic Regret with Adversarial Robustness in Partial Monitoring}

\begin{document}

\twocolumn[
\icmltitle{Exploration by Optimization with Hybrid Regularizers: \\
Logarithmic Regret with Adversarial Robustness in Partial Monitoring}

\begin{icmlauthorlist}
\icmlauthor{Taira Tsuchiya}{utokyo,riken}
\icmlauthor{Shinji Ito}{utokyo,riken,nec}
\icmlauthor{Junya Honda}{kyotou,riken}
\end{icmlauthorlist}

\icmlaffiliation{utokyo}{Department of Mathematical Informatics, The University of Tokyo, Tokyo, Japan}
\icmlaffiliation{nec}{NEC Corporation, Kanagawa, Japan (affiliation upon submission)}
\icmlaffiliation{riken}{RIKEN AIP, Tokyo, Japan}
\icmlaffiliation{kyotou}{Department of Systems Science, Kyoto University, Kyoto, Japan}

\icmlcorrespondingauthor{Taira Tsuchiya}{tsuchiya@mist.i.u-tokyo.ac.jp}

\icmlkeywords{online learning, partial monitoring, best-of-both-worlds, exploration-by-optimization, hybrid regularizer}

\vskip 0.3in
]



\printAffiliationsAndNotice{}  

\begin{abstract}
Partial monitoring is a generic framework of online decision-making problems with limited feedback. To make decisions from such limited feedback, it is necessary to find an appropriate distribution for exploration. Recently, a powerful approach for this purpose, \emph{exploration by optimization} (ExO), was proposed, which achieves optimal bounds in adversarial environments with follow-the-regularized-leader for a wide range of online decision-making problems. However, a naive application of ExO in stochastic environments significantly degrades regret bounds. To resolve this issue in locally observable games, we first establish a new framework and analysis for ExO with a hybrid regularizer. This development allows us to significantly improve existing regret bounds of best-of-both-worlds (BOBW) algorithms, which achieves nearly optimal bounds both in stochastic and adversarial environments. In particular, we derive a stochastic regret bound of $O(\sum_{a \neq a^*} k^2 m^2 \log T / \Delta_a)$, where $k$, $m$, and $T$ are the numbers of actions, observations and rounds, $a^*$ is an optimal action, and $\Delta_a$ is the suboptimality gap for action $a$. This bound is roughly $\Theta(k^2 \log T)$ times smaller than existing BOBW bounds. In addition, for globally observable games, we provide a new BOBW algorithm with the first $O(\log T)$ stochastic bound.
\end{abstract}

\section{Introduction}
\begin{table*}[t]
  \caption{Regret upper bounds for partial monitoring.
  Here the number of rounds is denoted by $T$, the number of actions by $k$, the number of outcomes by $d$, the maximum number of distinct symbols per action by $m$, the minimum suboptimality gap by $\Deltamin$, the game-dependent constant by $c_\calG$, and the corruption level by $C$.
  The variables $\calR^{\mathrm{loc}}$ and $\calR^{\mathrm{glo}}$ are the regret upper bounds of the proposed algorithms in stochastic environments for locally and globally observable games, respectively.
  TSPM is the bound by~\citet{Tsuchiya20analysis} for the linearized version of PM; refer to the paper for the definition of $\Lambda'$.
  ExpPM is by~\citet{lattimore20exploration}.
  PM-DEMD is by~\citet{Komiyama15PMDEMD}, where $D(\nu^*)$ is a distribution-dependent constant.
  ExpPM-BOBW is by~\citet{tsuchiya23best}.
  }
  \label{table:regret_PM}
  \centering
  \footnotesize
  \begin{tabular}{lllll}
    \toprule
    Observability
    & Algorithm & Stochastic & Adversarial & Corrupted \\
    \midrule
    \multirow{2}{6em}{Locally observable}
    &
    TSPM & $\displaystyle O\left( \frac{ k^2 m d \log T}{\Lambda'^2} \right)$  & --  & --  \\
    \rule{0pt}{3ex}
    &
    ExpPM & -- & $O\prn*{k^{3/2} m \sqrt{T \log k}}$ & -- \\
    \rule{0pt}{4ex}
    &
    ExpPM-BOBW &  $\displaystyle O\prn*{\frac{ k^4 m^2 \log T \log(k T)}{\Deltamin}}$ & $O\prn*{ k^{3/2} m \sqrt{T \log T \log k}}$   & $\calR^{\mathrm{loc}} + \sqrt{C \calR^{\mathrm{loc}}}$  
     \\
    \rule{0pt}{5ex}
    &
    \textbf{Ours (Theorem~\ref{thm:lob-main-theorem})} &  $\displaystyle O\prn[\Bigg]{ \sum_{a \neq a^*} \frac{k^2 m^2 \log T}{\Delta_a} }$ & $\displaystyle O\prn*{ k^{3/2} m \sqrt{T \log T}}$   & $\displaystyle \calR^{\mathrm{loc}} + \sqrt{C \calR^{\mathrm{loc}}}$  
     \\
    \midrule
    \multirow{2}{6em}{Globally observable}
    & PM-DMED & $O(D(\nu^*) \log T)$  & -- & -- \\
    \rule{0pt}{4ex}
    & ExpPM & -- & $O\prn*{(c_{\calG} T)^{2/3} (\log k)^{1/3}}$ & -- \\  
    \rule{0pt}{4ex}
    &
    ExpPM-BOBW & $\displaystyle O\left(\frac{c_{\mathcal{G}}^2 \log T \log(k T)}{\Deltamin^2}\right)$ & $O\prn*{ (c_{\mathcal{G}} T)^{2/3} (\log T \log(k T))^{1/3}}$ & $\calR^{\mathrm{glo}} + (C^2 \calR^{\mathrm{glo}})^{1/3}$     
    \\
    \rule{0pt}{4ex}
    &
    \textbf{Ours (Theorem~\ref{thm:gob-main-theorem})} & $\displaystyle O\left(\frac{c_{\mathcal{G}}^2 k \log T}{\Deltamin^2}\right)$ & $O\prn*{(c_{\mathcal{G}} T)^{2/3} (\log T)^{1/3}}$ & $\calR^{\mathrm{glo}} + (C^2 \calR^{\mathrm{glo}})^{1/3}$     
    \\
    \bottomrule
  \end{tabular}
\end{table*}

This paper investigates partial monitoring (PM), a comprehensive model for dealing with online decision-making problems with limited feedback~\citep{Rustichini99general, Piccolboni01FeedExp3}.
A PM game with $k$-actions and $d$-outcomes, denoted by $\calG = (\lossmat, \fbmat)$, is defined by a loss matrix $\lossmat \in [0,1]^{k \times d}$ and feedback matrix $\fbmat \in \Sigma^{k \times d}$, where $\Sigma$ is a set of feedback symbols.

This game unfolds in a sequential manner, involving interactions between a learner and an adversary for $T$ rounds.
At the start of the game, the learner and adversary observe~$\calL$ and $\Phi$.
At every round $t \in [T] \coloneqq \{1,\dots, T\}$, 
the adversary chooses an outcome $x_t \in [d]$. 
The learner then chooses an action $\at \in [k]$ without knowing $x_t$, incurs an unobserved loss $\lossmat_{\at x_t}$, and receives a feedback symbol~$\sigma_t = \fbmat_{\at x_t}$, where $A_{ij}$ is the $(i,j)$-element of matrix $A$.
The learner does not have access to outcome and loss, but only to feedback symbols.
The learner aims to minimize the cumulative loss over all the rounds with such limited feedback.

To measure the performance of the learner, we consider the (pseudo-)regret $\Reg_T$, which is defined as the difference between the cumulative loss of the learner and the single optimal action $a^*$, which is determined in hindsight.
More specifically, letting 
$
  \Reg_T(a)
  =
  \E \big[ \sumT \prn[]{\lossmat_{\at x_t} - \lossmat_{a x_t}} \big]
  ,
$
we define the regret by $\Reg_T = \Reg_T(a^*)$ for
optimal action
$
  a^* = \argmin_{a \in [k]} \E \big[ \sumT \calL_{a x_t} \big]
  .
$

The difficulty of PM games is determined based on its loss matrix and feedback matrix~\citep{Bartok11minimax, lattimore19cleaning}.
PM games fall into trivial, easy, hard, and hopeless games, for which its minimax regrets are~$0$, $\Theta(\sqrt{T})$, $\Theta(T^{2/3})$, and $\Theta(T)$, respectively. 
PM games for which the regret can be $O(\sqrt{T})$ and $O(T^{2/3})$ are called locally and globally observable, respectively.

PM has been mainly studied in stochastic environments and adversarial environments.
In stochastic environments, a sequence of outcomes $x_1, \dots, x_T \in [d]$ is drawn independently and identically from an unknown distribution $\nu^*$.
In contrast, in adversarial environments, outcomes are generated in an arbitrary manner and $x_t$ can depend on the past history of actions $(A_s)_{s=1}^{t-1}$.

For the adversarial environments, many algorithms have been developed to achieve the aforementioned minimax regrets and most of the algorithms are based on \emph{follow-the-regularized-leader} (FTRL) \citep{Piccolboni01FeedExp3, CesaBianchi06regret, lattimore20exploration}. 
In particular, \citet{lattimore20exploration} propose a strong approach called \emph{exploration by optimization} (ExO).
They prove that FTRL with the (negative) Shannon entropy (corresponding to the well-known Exp3 algorithm by~\citealt{freund97decision, auer2002nonstochastic}) combined with ExO achieves regret bounds of~$O(k^{3/2} m \sqrt{T \log k})$ 
for (non-degenerate) locally observable games and 
$O((c_{\calG} T)^{2/3} (\log k)^{1/3})$ 
for globally observable games.
Here, $m \leq \min\{|\Sigma|, d\}$ is the maximum number of distinct observations per action 
and $c_{\mathcal{G}}$ is a game-dependent constant defined in Section~\ref{sec:global}.
In the stochastic environments, there are also several algorithms; Starting from~\citet{bartok12CBP}, they include algorithms with optimal $O(\log T)$ regret upper bounds~\citep{Komiyama15PMDEMD, Tsuchiya20analysis}.

Although there are many algorithms for stochastic and adversarial environments, it is often the case in real-world problems that the underlying environment is unknown.
Hence, algorithms that achieve optimality in both stochastic and adversarial environments without knowing the underlying environment are desired. 
Such algorithms are called \emph{best-of-both-worlds} (BOBW) algorithms~\citep{bubeck2012best},
which have been widely investigated in recent years in a variety of online decision-making problems,~\eg~in \citet{gaillard2014second, wei2018more, zimmert2021tsallis}.
BOBW algorithms based on FTRL perform well also in stochastic environments with adversarial corruptions, an intermediate regime between stochastic and adversarial environments.
Extending the BOBW guarantee to this environment is standard and detailed in Appendix~\ref{sec:corrupted_env}.

The development of BOBW algorithms has recently extended from bandit problems to the PM problem~\citep{tsuchiya23best, tsuchiya23stability}.
\citet{tsuchiya23best} consider both locally observable and globally observable games.
For locally observable games, they construct a BOBW algorithm by extending the ExO approach.
As ExO was originally developed for adversarial environments, one could not derive a favorable regret lower bound that is needed for the \emph{self-bounding technique}~\citep{gaillard2014second, zimmert2021tsallis}, a standard technique for proving BOBW guarantees in FTRL (detailed in Section~\ref{sec:preliminaries}). 
To address this issue, the authors introduce \emph{a restriction of a feasible region} of action selection probability in ExO to derive a lower bound of the regret.
This technique allows us to provide an algorithm with a BOBW regret bound of~$O(k^{3/2} m \sqrt{T \log k \log T})$ in adversarial environments and
of $O({k^4 m^2 \log T \log (kT)}/{\Deltamin})$ for the minimum suboptimality gap $\Deltamin = \min_{a \neq a^*} \Delta_a$, where $\Delta_a = \E_{x_t \sim \nu^*} [ \calL_{a x_t} - \calL_{a^* x_t}  ]$ is the suboptimality gap for action~$a \in [k]$, in stochastic environments.

However, the bound for stochastic environments by their BOBW algorithm is significantly worse than the bounds solely for stochastic environments.
First of all, their bound scales with $\log^2 T$ rather than $\log T$, which seems to be due to the use of the Shannon entropy. In fact, there is no known algorithm that can achieve an $O(\log T)$ bound with the Shannon entropy (see \eg~\citealt{jin23improved}). 
Furthermore, their bound exhibits significant dependence on the number of actions $k$ and dependence on the smallest suboptimality gap $\Delta$ rather than action-wise suboptimality gap~$(\Delta_a)_{a \neq a^*}$.
(Due to space constraints, a further discussion of related work is deferred to Appendix~\ref{sec:additional-related-work}.)

\paragraph{Contributions of this Paper}
In order to improve these highly suboptimal dependencies on $k$, $T$, and $\Delta$, we introduce two major techniques to the PM problem.

The first is the introduction of FTRL with a hybrid regularizer consisting of the log-barrier and complement negative Shannon entropy.
In contrast to the Shannon entropy regularizer, 
the log-barrier regularizer is known to provide~$O(\log T)$ regret bounds~\citep{wei2018more,ito21parameter},
and combining it with the complement negative Shannon entropy~\citep{ito2021hybrid, ito22adversarially, tsuchiya23best} yields the regret bound in terms of action-wise suboptimality gaps $(\Delta_a)_{a \neq a^*}$,
in relatively simple settings such as multi-armed bandits and combinatorial semi-bandits.

This hybrid regularizer, however, cannot be employed for PM in a naive manner. 
This is because PM algorithms need to deal with complex feedback and suffer a very large regret without the ExO framework~\citep{lattimore20exploration},
and existing ExO analysis heavily relies on the specific properties of the Shannon entropy regularizer.
Hence, as a second technique, we analyze the optimal value of ExO when combined with the hybrid regularizer.
A new analysis to bound the optimal value allows us to optimize the ExO objective over the feasible region that is independent of the number of actions $k$. 
This improves the existing suboptimality in terms of $k$ (see Lemma~\ref{lem:stab_bound_pm_lb}).

Consequently, we significantly improve the regret upper bound in locally observable games.
In particular, we prove that the regret of the proposed algorithm is bounded by
\begin{equation}
  \Reg_T 
  = 
  O\prn*{
    \sum_{a \neq a^*} \frac{k^2 m^2 \log T}{\Delta_a}
  }
  \n 
\end{equation}
in stochastic environments and is bounded by~$\Reg_T = O(k^{3/2} m \sqrt{T \log T})$ in adversarial environments.
The bound in stochastic environments is roughly $\Theta(k^2 \log T)$ times smaller than the existing regret bound with the BOBW guarantee.
The bound for adversarial environments is~$O(\sqrt{\log k})$ times smaller than the existing regret bound with the BOBW guarantee. 
In addition, for globally observable games, we propose the first BOBW algorithm that achieves an $O(\log T)$ regret upper bound in stochastic environments by making a sacrifice of dependence on $k$.

A comparison of our bounds with existing regret bounds in PM is summarized in Table~\ref{table:regret_PM}.
Note that comparisons with existing algorithms designed exclusively for stochastic environments, TSPM and PM-DMED in Table~\ref{table:regret_PM}, are difficult for the following reasons.
The upper bound of TSPM is for linearized variants of PM, called linear PM.
The upper bound of PM-DMED should be better than our algorithm as it is asymptotically optimal, but $D(\nu^*)$ is expressed by a complex optimization problem, and its dependencies on $k$, $m$, and $d$ are unknown.

\section{Preliminaries}\label{sec:preliminaries}

This section provides basic concepts for PM
and introduces FTRL, the core framework of proposed algorithms. 

\paragraph{Notation}
Let $e_{i} \in \{0,1\}^k$ be the $i$-th standard basis in $\R^k$ and $\ones$ be the all-one vector.
A scalar $x_i$ is the $i$-th element of vector $x$, and $A_{ij}$ is the $(i,j)$-element of matrix $A$.
For vector $x$, let $\nrm{x}_p$ for $p \in [1, \infty]$ be the $\ell_p$-norm and write $\nrm{x} = \nrm{x}_2$.
For matrix $A$, we let $\nrm{A}_\infty = \max_{i,j} \abs{A_{ij}}$ be the maximum norm.
The $(k-1)$-dimensional probability simplex is denoted by $\calP_{k} = \{ p \in [0,1]^k \colon \nrm{p}_1 = 1 \}$.
Let $\mathrm{int}(A)$ be the interior of set $A$ and $\dom{\psi}$ be the domain of function $\psi$.
Given differentiable convex function $\psi$, the Bregman divergence induced by $\psi$ from $y$ to $x$ is defined by
$D_{\psi}(x, y) = \psi(x) - \psi(y) - \innerprod{\nabla \psi(y)}{x - y}$,
and
the convex conjugate of $\psi$ by
$\psi^*(x) = \sup_{a \in \R^k} \set{\innerprod{x}{a} - \psi(a)}$.
Table~\ref{table:notation} in Appendix~\ref{sec:notation} summarizes the notation used in this paper.

We assume that the optimal action $a^*$ is unique in stochastic environments, and thus $\Delta_a > 0$ for all $a \neq a^*$.
This assumption is standard in the literature to develop BOBW algorithms~\citep{gaillard2014second,wei2018more,zimmert2021tsallis,tsuchiya23best}.

\subsection{Basic Concepts in Partial Monitoring}
Here, we introduce basic concepts used in PM.
We consider a PM game $\calG = (\lossmat, \fbmat)$,
and denote the maximum number of distinct symbols in a single row of $\fbmat \in \Sigma^{k \times d}$ over all rows by $m \le |\Sigma|$.
We use $\ell_a = e_a^\top \calL \in \R^d$ to denote the $a$-th row of $\mathcal{L}$.

Two different actions $a$ and $b$ are \emph{duplicate} if $\ell_a = \ell_b$.
We can decompose possible distributions of $d$ outcomes in $\calP_d$ based on the loss matrix: 
For every action $a\in[k]$, \emph{cell} is the set of probability vectors in $\calP_d$ for which action $a$ is optimal, that is,
$\calC_a = \crl{u \in \calP_{d} \colon \max_{b\in[k]} (\ell_a - \ell_b)^\top u \le 0}$.
Each cell is a convex closed polytope.
We denote the dimension of the affine hull of $\calC_a$ by $\dim(\calC_a)$.
Action $a$ is \emph{dominated} if $\calC_a = \emptyset$.
  Non-dominated action $a$ is \emph{Pareto optimal} if $\dim(\calC_a) = d-1$ and \emph{degenerate} if $\dim(\calC_a) < d-1$.
We denote the set of Pareto optimal actions by $\Pi$.
Two Pareto optimal actions $a, b \in \Pi$ are \emph{neighbors} if $\dim(\calC_a \cap \calC_b) = d-2$.
The notion of neighboring will be used to classify PM games based on their difficulty in Section~\ref{susubsec:observability}.
Using this neighborhood relation, we can construct the undirected graph that is known to be connected (\citealt[Lemma 37.7]{lattimore2020book}).
This fact will be used for loss difference estimations between Pareto optimal actions in Section~\ref{subsubsec:loss_diff_est}.

A PM game is \emph{non-degenerate} if the game does not contain degenerate actions.
From hereon, we assume that PM game~$\calG$ is non-degenerate and contains no duplicate actions.

\subsubsection{Observability}\label{susubsec:observability}
The above notion of neighborhood relations is used to define \emph{observability} that characterizes the difficulty of PM games.
\begin{definition}
Neighbouring actions $a$ and $b$ are \emph{globally observable} if there exists $w_{e(a,b)} \colon [k] \times \Sigma \to \R$ such that
\begin{equation}\label{eq:observability}
  \sum_{c=1}^k w_{e(a,b)}(c, \fbmat_{cx})
  = 
  \lossmat_{ax} - \lossmat_{bx} 
  \;\ \text{for all}\; x \in [d]
  \com 
\end{equation}
where $e(a,b) = \{a,b\}$.\footnote{Here, the set $\{a, b\}$ and the edge $e(a,b)$ connecting $a$ and $b$ in an undirected graph generated by the neighborhood relations are regarded as equivalent. We will omit actions $a$ and $b$ connecting the edge when it is clear from the context.} 
Neighbouring actions $a$ and $b$ are \emph{locally observable} if there exists $w_{e}$ satisfying~\eqref{eq:observability} and $w_{e(a,b)}(c,\sigma) = 0$ for all $c \not\in \{a,b\}$ and $\sigma \in \Sigma$.
A PM game is called globally (resp.~locally) observable if all neighboring actions are globally (resp.~locally) observable.
\end{definition}
Locally observable games are globally observable,
In the remaining sections, we assume that $\calG$ is globally observable.

\subsubsection{Estimating Loss Differences}\label{subsubsec:loss_diff_est}
Using the definitions up to this point, we explain how to estimate losses in PM.
In PM it is common to estimate the loss difference between actions rather than estimating the loss itself.
Let $\calH$ be the set of all functions from $[k]\times\Sigma$ to $\R^k$.
Let $\mathscr{T}$ be an in-tree over $\Pi$ induced by the neighborhood relations with an arbitrarily chosen root. 
Then, 
from the fact that the undirected graph induced by $\Pi$ is connected,
it follows that loss differences between Pareto optimal actions can be estimated as follows.
\begin{lemma}[{\citealt[Lemma 4]{lattimore20exploration}}]\label{lem:Gdiff_Ldiff}
For any globally observable game, there exists a function $G \in \calH$ such that for any Pareto optimal actions $a, b \in \Pi$,
\begin{equation}
  \!\!\!
  \forall 
  x \in [d] \,,\,
  \sum_{c=1}^k (G(c, \fbmat_{cx})_a - G(c, \fbmat_{cx})_b) 
  \!=\!
  \calL_{ax} - \calL_{bx}
  \per
  \label{eq:Gdiff_Ldiff}
\end{equation}
In particular, the following $G^{\circ}$ satisfies~\eqref{eq:Gdiff_Ldiff}:
\begin{equation}\label{eq:G_0}
  G^{\circ}(a, \sigma)_b 
  = 
  \sum_{e \in \mathrm{path}_\mathscr{T}(b)} w_e(a, \sigma)
  \;
  \mbox{for}
  \;
  a \in \Pi
  \com
\end{equation}
where 
$\mathrm{path}_\mathscr{T}(b)$ is the set of edges from $b \in \Pi$ to the root on $\mathscr{T}$.
\end{lemma}

\subsubsection{Examples of partial monitoring games}
For an intuitive understanding of the concept introduced above, we present typical examples of PM games.
\paragraph{Multi-armed bandits}
One of the most well-known examples of PM games is multi-armed bandits with finite loss support~\cite{LaiRobbins85,auer2002nonstochastic}.
For example, the $k$-armed Bernoulli bandits can be represented by ${k \times 2^k}$ loss and feedback matrices.
In particular when~$k = 2$, 
\begin{equation}
  \calL = 
  \begin{pmatrix}
    0 & 1 & 0 & 1 \\
    0 & 0 & 1 & 1 \\
  \end{pmatrix}
  \com 
  \;\;
  \Phi = 
  \begin{pmatrix}
    0 & 1 & 0 & 1 \\
    0 & 0 & 1 & 1 \\
  \end{pmatrix}
  \per 
  \n
\end{equation}
One can see that in this case, we have $m = 2$ and the game has no duplicate actions and is non-degenerate.

\paragraph{Matching pennies}
Next, we consider the problem known as matching pennies, which is useful for understanding PM games~\cite{lattimore2020book,lattimore20exploration}.
In this setting, the learner needs to classify emails sequentially delivered to a mailbox are positive $\mathsf{P}$ (spam) or negative $\mathsf{N}$ (ham).
The learner has three choices of actions: (1) classify an email as $\mathsf{P}$, (2) classify an email as $\mathsf{N}$, or (3) let a human inspect an email and determine whether it is $\mathsf{P}$ or $\mathsf{N}$.
Assume that the learner can only confirm if it is spam or ham by choosing the third action and that the inspection cost by a human is~$c_q \geq 0$.
When $\mathsf{P}$ is wrongly classified as $\mathsf{N}$, the learner suffers a loss of $c_{\mathsf{P} \to \mathsf{N}} > 0 $ for misclassification and vice versa.
The goal is to choose a sequence of actions so that the cumulative cost is minimized.

This problem can be indeed formulated as a PM game with~$k = 3$ and $d = 2$ with
\begin{equation}
  \calL
  =
  \begin{pmatrix}
    0 & c_{\mathsf{N} \to \mathsf{P}} \\
    c_{\mathsf{P} \to \mathsf{N}} & 0 \\
    c_q & c_q
   \end{pmatrix}
  \com 
  \;\;
  \Phi
  =
  \begin{pmatrix}
    \mathsf{None} & \mathsf{None} \\
    \mathsf{None} & \mathsf{None} \\
    \mathsf{P} & \mathsf{N} \\
  \end{pmatrix}
  \per
  \n
\end{equation}
It is known that this problem falls into easy or hard games depending on the parameters in $\lossmat$.
The most well-studied instance is with $c_{\mathsf{N} \to \mathsf{P}} = c_{\mathsf{P} \to \mathsf{N}} = 1$,
and the problem becomes non-degenerate and locally observable if $c_q \in (0,1/2)$ and globally observable if $c_q > 1/2$, and trivial if $c_q = 0$~\citep{lattimore20exploration}.

\subsection{Follow the Regularized Leader}
We rely on follow-the-regularized-leader (FTRL) to develop the proposed algorithms.
In FTRL, we compute a probability vector $q_t \in \calP_k$ that minimizes the sum of expected cumulative loss and convex regularizer $\psi_t \colon \calP_k \to \R$ as follows:
\begin{equation}\label{eq:def_q}
  \qt
  =
  \argmin_{q \in \calP_{\Pi}}
  \,
  \tri[\Bigg]{\sum_{s=1}^{t-1} \hat y_s,\, q}
  +
  \psi_t(q)
  \com 
\end{equation}
where
the set $\calP_{\Pi} \coloneqq \{p \in \calP_k \colon  p_a = 0, \forall a \not\in \Pi \}$ is the convex closed polytope on the probability simplex with nonzero elements only at indices in $\Pi$, and $\hat y_t \in \R^k$ is an estimator of the loss (difference) at round $t$.
The following lemma is well-known and used when bounding the regret of FTRL.
\begin{lemma}\label{lem:ftrl}
Let $q_1, \dots, q_T$ be determined by~\eqref{eq:def_q}.
Then, for any $u \in \calP_k$, $\sumT \innerprod{\hat{y}_t}{q_t - u}$ is bounded from above by
\begin{align}\label{eq:lem-ftrl}
  &
  \sumT
  \prn[\big]{
  \psi_t(q_{t+1})
  -
  \psi_{t+1}(q_{t+1})
  }
  +
  \psi_{T+1} (u) 
  -
  \psi_1 (q_1)
 \nn
 &\quad
  +
  \sumT
  \prn[\big]{
  \innerprod{\hat{y}_t}{q_t - q_{t+1}}
  -
  D_{\psi_t}(q_{t+1}, q_t)
  }
  \per
\end{align}
\end{lemma}
We refer to the first and second lines in~\eqref{eq:lem-ftrl} as \emph{penalty} and \emph{stability} terms.
The proof of this lemma is standard in the literature and can be found \eg~in \citet[Chapter 28]{lattimore2020book} and \citet[Chapter 7]{orabona2019modern}.
We include the proof for completeness in Appendix~\ref{sec:proof-common}.

To prove the BOBW guarantee with FTRL, it is essential to rely on the self-bounding technique~\citep{gaillard2014second, zimmert2021tsallis}.
This approach combines upper and lower bounds that depend on FTRL output $(q_t)_t$ (or action selection probability $(p_t)_t$).
For example, if we have $\Reg_T \leq c_{\mathsf{U}}  \sqrt{ Q \log T} $ and $\Reg_T \geq c_{\mathsf{L}} Q$ for $Q = \E\brk[\big]{ \sumT \sum_{a \neq a^*} q_{ta} }$ hold, then from the inequality $bx - ax^2 \leq b^2 / (4a)$ for $a, b > 0$,
the regret is bounded as
\begin{equation}
  \Reg_T 
  =
  2 \Reg_T - \Reg_T
  \leq 
  2 c_{\mathsf{U}} \sqrt{Q \log T} - c_{\mathsf{L}} Q
  \leq 
  \frac{c_{\mathsf{U}}^2 \log T}{c_{\mathsf{L}}}
  \per 
  \nn 
\end{equation}
Note that the smaller $c_{\mathsf{L}}$ becomes the larger the regret upper bound becomes, and we will see that deriving the tight \emph{lower} bound contributes to improving the dependence on~$k$.

In the problem of estimating losses from indirect feedback like PM, it is common to apply some transformation to FTRL output $q_t$ to reduce the variance of the loss estimator, instead of using $q_t$ as action selection probability $\pt$.
In the next section, we will investigate a favorable transformation.

\section{Exploration by Optimization}\label{sec:exo}
This section first reviews the existing techniques of exploration by optimization (ExO),
and then proposes and analyzes a novel ExO with a hybrid regularizer, which forms the key basis for our algorithm and regret analysis in locally observable games.

\subsection{Principles of Exploration by Optimization}\label{subsec:exo_orignal}

ExO determines action selection probability $p_t$ from FTRL output $q_t$~\citep{lattimore20exploration}.
In PM games, constructing a loss estimator directly from the output of FTRL tends to result in large variances. 
This issue arises because $q_t$ in~\eqref{eq:def_q} does not take into account the amount of information provided by each action. 
It is common to add a certain amount of exploration to $q_t$, for example by mixing a uniform distribution with $q_t$ when computing $p_t$.
Instead, ExO determines $p_t$ from $q_t$ by minimizing a component of a regret upper bound consisting of the transformation term~$(\pt - \qt)^\top \calL e_{x_t}$ and the stability term in~\eqref{eq:lem-ftrl}, which is linked to the variance of a loss estimator. 

It is well-known that the stability term is bounded as follows (see \eg~\citealt[Lemma 26]{lattimore21mirror}).
For a strongly convex function $\phi \colon \R \to \R$ and any $p, q \in (0,1)$ with $q \in \interior(\dom{\phi})$, $\ell \in \R$, and $z \in \R$, it holds that
\begin{equation}
  \prn{q - p}{z} - D_{\phi}(p, q)
  \!\leq\!
  D_{\phi^*}(\nabla \phi(q) - z, \nabla \phi(q))
  \eqqcolon
  \stab^\phi_q(z)
  \per 
  \label{eq:def_calS}
\end{equation}
Note that $\stab^\phi_q(z)$ is convex with respect to $z$.

We consider FTRL with regularizer 
\begin{equation}
  \psi(q) = \sumak \beta_a \phi(q_a)
  \label{eq:reg_arm_wise}
\end{equation}
with action-dependent learning rate $\beta \in \R_+^k$ and strongly convex function $\phi \colon \R \to \R$.
Let $\calH^{\circ}$ be the set of $G \colon [k] \times [d] \to \R^k$ satisfying~\eqref{eq:Gdiff_Ldiff}, that is,
\begin{align}
  \calH^{\circ}
  =
  \Bigg\{
    G \colon
    \sum_{c=1}^k & \prx{ G(c, \Phi_{cx})_a - G(c, \Phi_{cx})_b }
    = 
    \calL_{bx} - \calL_{cx}  \
   \nn 
   &\quad\mbox{ for all } a, b \in \Pi \mbox{ and } x \in [d]
  \Bigg\}
  \per 
  \n
\end{align}
Then the objective function of the optimization problem over $p \in \calP_k$ and $G \in \calH^{\circ}$ in ExO with the regularizer $\psi$ in~\eqref{eq:reg_arm_wise} is
\begin{align}\label{eq:exo-obj}
  &\objexo(p, G ; q, (\beta, \phi )) 
  = 
 \nn
 &
  \max_{x \in [d]} 
  \Bigg[
  {(p-q)^\top  \lossmat e_x}
  +
  \sum_{b=1}^k \beta_b 
  \sumak
  p_a \stab_{q_b}^{\phi}  \prn*{ \frac{ G(a, \Phi_{ax})_b}{ \beta_b p_a}  } 
  \Bigg] 
  \per 
\end{align}
Note that the first and second terms correspond to the transformation and stability terms, respectively.
Note also that~\eqref{eq:exo-obj} is convex with respect to $p$ and $G$ since $\stab_{q_a}^{\phi}$ is convex and hence its perspective $(p,g) \mapsto p \, \stab_q^\phi(g/p)$ is also convex~\citep[Chapter 3]{boyd04convex}.

\begin{remark}
The objective function of ExO in~\eqref{eq:exo-obj} is a slightly generalized version of the existing ExO in the use of the \emph{action-wise learning rate} $\beta \in \Rp^k$,
while existing formulations use the same learning rate for each action~\citep{lattimore20exploration, lattimore21mirror, tsuchiya23best}.
This generalization enables us to obtain a $\Delta_a$-wise regret upper bound. 
\end{remark}

\paragraph{Original exploration by optimization}
The original formulation of ExO by~\citet{lattimore20exploration} employs the (negative) Shannon entropy regularizer $\psi_t(q) = \frac{1}{\eta} \sumak \phiNS(q_a) = \frac{1}{\eta} \sumak q_a \log q_a $ and solve the following optimization problem:
\begin{equation}\label{eq:exo-original}
  (p, G)
  =
  \argmin_{p \in \calP_k,  \, G \in \calH^{\circ}}
  \objexo(p, G; q, (\ones/\eta, \phiNS) )
  \per 
\end{equation}
Note that the feasible region for action selection probability $p$ here is the probability simplex $\calP_k$.
They prove that the optimal value of~\eqref{eq:exo-original} is bounded by $3m^2 k^3 \eta $. 
Roughly speaking, letting 
\eqref{eq:exo-obj}
be $\stabtrans(p; e_x)$
they construct~$p^{\mathsf{NS}}(\nu) \in \calP_k$ to prove
\begin{align}
  \min_{p \in \calP_k} 
  \objexo(p)
  &
  \leq 
  \max_{\nu \in \calP_d} \min_{p \in \calP_k} 
  \stabtrans(p; \nu)
  \nn
  &\leq 
  \max_{\nu\in\calP_d} \stabtrans(p^{\mathsf{NS}} ; \nu)
  \leq
  3 m^2 k^3 \eta 
  \com 
  \label{eq:exo-arg}
\end{align}
where the first inequality follows from Sion's minimax theorem.\footnote{We abuse the notation and ignore several variables.}

\paragraph{Exploration by optimization for best-of-both-worlds guarantees}
Since the original ExO in~\eqref{eq:exo-original} considers only adversarial environments,
the self-bounding technique, which requires a lower bound in terms of FTRL output $q$, 
is not applicable. 
In particular, there is a case where the solution of~\eqref{eq:exo-original} satisfies $p_a = 0$ even when $q_a > 0$.
To address this issue, the following optimization problem
over the restricted feasible region $\calP'(q) \subset \calP_k$ was investigated~\citep{tsuchiya23best}:
\begin{equation}\label{eq:exo-bobwpm}
\begin{split}
  ( p, G )
  &=
  \argmin_{p \in \calP'(q), \, G \in \calH^{\circ}}
  \objexo(p, G; q, (\eta, \phiNS))
  \quad \mbox{for} \;
  \\ 
  \calP'(q) &= \{ p \in \calP_k \colon p_a \geq q_a / (2k) \mbox{ for all } a \in [k] \}
  \per
\end{split}
\end{equation}
Note that the feasible region for action selection probability $p$ is \emph{truncated} probability simplex $\calP'(q)$, which allows to use the self-bounding technique since $p \geq q / (2k)$.
Despite the fact that minimization problem~\eqref{eq:exo-bobwpm} restricts the feasible region, the optimal value in~\eqref{eq:exo-bobwpm} is bounded by $3 m^2 k^3 \eta$ as in~\eqref{eq:exo-original}.
This can be confirmed by observing~$p^{\mathsf{NS}} \in \calP'(q)$.

The regret upper bound for stochastic environments in~\citet{tsuchiya23best, tsuchiya23stability} obtained by solving the optimization problem~\eqref{eq:exo-bobwpm} is $O(k^4 m^2 \log T \log(kT) / \Delta)$.
Although this bound is poly-logarithmic in $T$,
it is highly suboptimal.
First, the dependence on $T$ is $O(\log^2 T)$, which is suboptimal and comes from the use of the Shannon entropy regularizer.
In fact, all but one regret upper bounds based on the Shannon entropy regularizer are at least~$O(\log^2 T)$ even for simple settings like multi-armed bandits, motivating us to use other regularizers.\footnote{The only exception, a reduction approach by~\citet{dann23blackbox} is difficult to use in PM. See Section~\ref{sec:local} for a detailed discussion.}
The bound also depends on the minimum suboptimality gap $\Delta$ rather than action-wise suboptimality gaps $(\Delta_a)_{a \neq a^*}$.
Furthermore, the dependence on $k$ is $k^4$, which is highly suboptimal and comes from the use of probability simplex $\calP'_k(q)$ in~\eqref{eq:exo-bobwpm}.
In order to overcome these issues, 
we will employ a hybrid regularizer as the main regularizer, which is known to achieve the~$O(\log T)$ and $(\Delta_a)_{a}$-dependent regret bounds in simpler online settings.
We then investigate its behavior under the ExO technique.

\subsection{Exploration by Optimization with Hybrid Regularizer}\label{subsec:exo_hybrid}
Here, we present a novel algorithm and analysis of ExO with the hybrid regularizer to derive an improved regret bound in Section~\ref{sec:local}.
Let 
\begin{align}\label{eq:lb_and_cns}
  \phiLB(x) = - \log x
  \com \,
  \phiCNS(x) = (1-x) \log (1-x)
\end{align}
be the component of the negative Shannon entropy and log-barrier regularizer, respectively.
We then consider the regularizer $\psi \colon \calP_k \to \R_+$ defined by
\begin{equation}\label{eq:defpsiexo}
\begin{split}
  \psi(p) 
  &= 
  \sumak \beta_{a} \phi(p_a)
  \com
  \quad\mbox{with}\quad
 \\
  \phi(x) 
  &= 
  \prn[\big]{ \phiLB(x) + x - 1} + \gamma \prn[\big]{ \phiCNS(x) + x} 
\end{split}
\end{equation}
for $\gamma = \log T$ and vector $\beta \in \R_+^k$.
Note that $\psi_t \geq 0$ since $\phiLB(x) + x - 1 \geq 0$  for all $x > 0$ and $\phiCNS(x) + x \geq 0$ for all $x \leq 1$.
Let
\begin{equation}\label{eq:half-truncated-simplex}
  \calR(q) = \{ p \in \calP_k \colon p_a \geq q_a / 2 \mbox{ for all } a \in [k] \}
\end{equation}
be the probability simplex truncated to have more than half of the probability vector $q$. 
We then consider the optimization problem of ExO objective in~\eqref{eq:exo-obj} over this set as follows:
\begin{equation}
  ( p, G )
  =
  \argmin_{p \in \calR(q), \, G \in \calH^{\circ} }
  \objexo(p, G; q, (\beta, \phi) )
  \label{eq:exo}
\end{equation}
with $\phi$ in~\eqref{eq:defpsiexo}.
Let $\OPT_q(\beta)$ be the optimal value of~\eqref{eq:exo}. 
\begin{remark}
Recall that~\eqref{eq:exo-bobwpm} considers the truncation of the feasible region that depends on $k$, which makes the dependence on $k$ in the stochastic regret bound worse.
In contrast, $\calR(q)$ does not depend on $k$, which will improve the dependence on $k$ in the regret bound as we will see in Section~\ref{sec:local}.
Note that for this change of the truncation,
several techniques are needed as detailed in the following.
\end{remark}
The optimal value $\OPT_q(\beta)$ is bounded as follows:
\begin{lemma}\label{lem:stab_bound_pm_lb}
Consider non-degenerate locally observable partial monitoring games.
Suppose that $\gamma \geq 2$ and $\beta \geq 4 m k$.
Then the optimal value $\OPT_q(\eta)$ of the exploration by optimization in~\eqref{eq:exo} with regularizer~\eqref{eq:defpsiexo} is bounded as
\begin{equation}\label{eq:opt_bound}
  \OPT_q(\beta)
  \leq 
  2 m^2 k^2
  \sum_{b=1}^k
  \frac{1}{\beta_b}
  \min  
  \brx{
    q_b
    ,
    \frac{1 - q_b}{\gamma q_b}
  }  
  \per  
\end{equation}
\end{lemma}
The proof of Lemma~\ref{lem:stab_bound_pm_lb} can be found in Appendix~\ref{sec:proof-stab-wto}.
The first benefit of this lemma, compared to existing ones, is that it establishes a bound for the optimization problem over $\calR(q)$ (in~\eqref{eq:exo}), a feasible region independent of $k$.
This approach leads to an improvement in the final regret upper bound by a factor of $k$ (see also Remark~\ref{rem:shannon_improve_k}), compared to ExO in~\eqref{eq:exo-bobwpm}.
The second benefit is that the obtained upper bound depends on FTRL output $q$, whereas the existing bound on optimal values of ExO is independent of $q$.
This distinction is crucial when applying the self-bounding technique with regularizer in~\eqref{eq:defpsiexo}.
When applying the self-bounding technique, it is essential for $q$ to appear in the upper bound of the stability or penalty term, where the objective of ExO consists of the stability and transformation terms.
If we use the log-barrier regularizer then the penalty term is no longer dependent on $q$ unlike when using the Shannon entropy regularizer, but the self-bounding technique is applicable due to Lemma~\ref{lem:stab_bound_pm_lb}, which bounds the stability term (and transformation term) in terms of $q$.

A crux of the proof is to consider feasible solution $p^{\mathsf{H}} = (q + W_{\nu}(q)) / 2$ to upper-bound the optimal value of ExO, where $W_{\nu}(\cdot)$ is called a \emph{water transfer operator}.
Here, the water transfer operator $W_{\nu}$ transforms FTRL output $q$ into a probability vector that has useful properties in locally observable games (see Lemma~\ref{lem:wto} in Appendix~\ref{sec:proof-stab-wto} and \citealt{lattimore2019information} for details).
This feasible solution $p^{\mathsf{H}}$ clearly satisfies $p^{\mathsf{H}} \geq q/2$, that is, $p^{\mathsf{H}} \in \calR(q)$.
It remains to show that $\Omega(p^{\mathsf{H}}, G^\circ)$ is bounded by the RHS of~\eqref{eq:opt_bound}.
Note that we cannot use the existing analysis mixing uniform distribution\footnote{This in fact is necessary with the Shannon entropy when bounding the stability term.} $\ones / k \in \calP_k$ to construct $p^\mathsf{NS} = (1-\gamma) W_{\nu}(q) + \gamma \ones / k$ for $\gamma \in (0,1/2]$ instead of $p^{\mathsf{H}}$.
This is because $p^{\mathsf{NS}}(\nu)$ is not only outside of our feasible region $\calR(q)$ but also results in an upper bound that does not depend on $q$, due to the regret coming from the transformation term.
To deal with this issue, the proof of Lemma~\ref{lem:stab_bound_pm_lb} introduces an analysis that bounds the stability term of the log-barrier regularizer without mixing uniform distribution by exploiting the property of the water transfer operator.

\begin{remark}\label{rem:shannon_improve_k}
By mimicking the construction of $p^\mathsf{H}$ in the proof of Lemma~\ref{lem:stab_bound_pm_lb}, we can improve the regret bound in the stochastic environment in~\citet{tsuchiya23best, tsuchiya23stability} by a factor of $k$.
In~\citet{lattimore20exploration, tsuchiya23best, tsuchiya23stability}, $p^\mathsf{NS}$ is constructed as $p^\mathsf{NS} = (1-\gamma) W_{\nu}(q) + \gamma \ones / k$ with a certain $\gamma \in (0,1/2]$.
Instead, by introducing $q/2$ so that $p^\mathsf{NS} = q/2 + ((1-\gamma) W_{\nu}(q) + \gamma \ones / k) /2 $, we can improve the regret upper bounds in~\citet{tsuchiya23best, tsuchiya23stability} in the stochastic environments by a factor of $k$.
Still, just solely applying this improvement does not lead to the regret bound that we will obtain in this paper.
\end{remark}

\section{Locally Observable Games}\label{sec:local}
This section proposes an improved BOBW algorithm for locally observable games and derives its regret bounds based on the analysis for ExO with the hybrid regularizer provided in Section~\ref{sec:exo}.

\begin{algorithm}[t]
  \caption{
  Algorithm for locally observable games
  }
  \label{alg:local}
  \begin{algorithmic}
  \FOR{$t = 1, 2, \ldots$} 
  \STATE Compute $\qt$ via FTRL~\eqref{eq:def_q} with 
  regularizer~\eqref{eq:defpsilocal}.
  
  \STATE Compute 
  $\pt$ and $G_t$ by exploration by optimization in~\eqref{eq:exo} with $q \leftarrow q_t$, $\phi \leftarrow \phi$ in~\eqref{eq:defpsilocal}, and $\beta \leftarrow \beta_t$ in~\eqref{eq:def_lr_local}.
  
  \STATE Sample $\at \sim \pt$, observe $\Phi_{A_t x_t} \in \Sigma$, and compute $\displaystyle \hat{y}_t = {G_t(A_t, \sigma_t)}/{\ptAt}$.
  \ENDFOR
  \end{algorithmic}
  \end{algorithm}

\subsection{Proposed Algorithm}
The proposed algorithm is based on ExO developed in Section~\ref{subsec:exo_hybrid}.
We use the regularizer~\eqref{eq:defpsiexo} for FTRL in Section~\ref{sec:exo}:
The regularizer $\psi_t \colon \calP_k \to \R_+$ in~\eqref{eq:def_q} is defined by
\begin{equation}\label{eq:defpsilocal}
  \begin{split}
  \psi_t(q) 
  &= 
  \sumak \beta_{ta} \phi(q_a) \com
  \\
  \phi(x) 
  &= 
  \prn{ \phiLB(x) + x - 1} + \gamma \prn{ \phiCNS(x) + x} \com
  \end{split}
\end{equation}
where 
$\phiLB$ and $\phiCNS$ are defined in~\eqref{eq:lb_and_cns}.
Note that this regularizer corresponds to that in~\eqref{eq:defpsiexo}.
The learning rate $\beta_{t} = (\beta_{t1}, \dots, \beta_{tk})^\top \in \Rp^k$ in~\eqref{eq:defpsilocal} are set to
\begin{align}
  \beta_{ta} &= \max\{4 m k, \beta'_{ta}\}
  \com \quad
  \beta'_{ta} = c \sqrt{\alpha_0 + \frac{1}{\gamma} \sum_{s=1}^{t-1} \alpha_{sa}}
  \com  \nn 
  \alpha_{ta} &= \min\brx{q_{ta}, \frac{1-q_{ta}}{\gamma q_{ta}}} \com 
  \label{eq:def_lr_local}
\end{align}
where $\alpha_0 = \frac{1}{\gamma^{3/2} } + \epsilon$ for $\epsilon > 0$ and $c = 2 m k$.

We compute $p_t$ and $G_t$ by~\eqref{eq:exo} with $q \leftarrow q_t$ and $\beta \leftarrow \beta_t$.
Using this $G_t$, we estimate the loss at round $t$ by 
$
  \yhat_t = {G_t(A_t, \sigma_t)}/{p_{tA_t}} 
  .
$
The proposed algorithm for locally observable games is summarized in Algorithm~\ref{alg:local}.

\subsection{Regret Bounds and Analysis}
Here we provide the regret bound and analysis of the proposed algorithm.
The proposed algorithm achieves the following bound:
\begin{theorem}\label{thm:lob-main-theorem}
Consider any non-degenerate partial monitoring games with horizon $T \geq 8$.
In stochastic environments, Algorithm~\ref{alg:local} with $c = 2mk$ achieves
\begin{equation}
  \Reg_T 
  = 
  O \prn[\Bigg]{
    \sum_{a \neq a^*}
    \frac{m^2 k^2 \log T}{\Delta_a}
  }  
  \per
  \label{eq:regret_local_stoc}
\end{equation}
In adversarial environments, the same algorithm achieves
\begin{equation}
  \Reg_T =
  O \prx{
    k^{3/2} m \sqrt{T \log T}
    +
    k^2 m \log T
    +
    k^3 m
  }  
  \per 
  \n 
\end{equation}
Further, in stochastic environments with adversarial corruptions, the same algorithm achieves $\Reg_T = O(\calR^{\mathrm{loc}} + \sqrt{C \calR^{\mathrm{loc}}})$ with $\calR^{\mathrm{loc}}$ denoting the RHS of~\eqref{eq:regret_local_stoc}.
\end{theorem}
The above bounds are significantly better than the existing bounds for PM in~\citet{tsuchiya23best}.
The bound for the stochastic environments depends on $\log T$ rather than $\log^2 T$, $(\Delta_a)_{a\neq a^*}$ instead of $\Delta_{\min}$, and $k^3$ instead of $k^4$.
The bound for the adversarial environment is $O(\sqrt{\log T})$ times smaller than the bound in~\citet{tsuchiya23best}.

\begin{remark}
Recently, \citet{dann23blackbox} proposed a framework for converting a bandit algorithm in adversarial environments into a BOBW algorithm with an $O(\log T)$ upper bound in stochastic environments in a blackbox manner. 
However, it is unclear whether this framework can be applied to PM. 
This is because, in order to use this framework, losses of selected actions must be observable, which does not apply to PM.
They consider how to apply their framework to online learning with feedback graphs (a.k.a.~graph bandits), a setting in which the selected action is not always observed.
However, this approach requires constructing a \emph{surrogate} loss for the problem, and it is highly unclear whether such a construction is possible for PM.
\end{remark}

In the following, we sketch the proof of Theorem~\ref{thm:lob-main-theorem}.
Define
$
  u
  = 
  \left(1-\frac{k}{T}\right) e_{a^*} + \frac{1}{T}
  .
$
Using this, we start by decomposing the regret as follows.
\begin{lemma}\label{lem:lob_1}
The regret of Algorithm~\ref{alg:local} is bounded from above by
\begin{align}\label{eq:lob_1}
  &
  \E\left[
    \sumT
    \prn[\big]{
    \psi_{t}(q_{t+1})
    -
    \psi_{t+1}(q_{t+1})
    }
    +
    \psi_{T+1} (u) 
    -
    \psi_1 (q_1)
  \right]
 \nn
 &\quad
  +
  \E\Bigg[
    \sumT
    \max \brx{0, \OPT_{q_t}(\beta_t)}
  \Bigg]
  +
  k
  \per
\end{align}
\end{lemma}
Note that the first term in~\eqref{eq:lob_1} corresponds to the penalty term.
The proof of this lemma is similar to Lemma 7 of~\citet{tsuchiya23best}.
The RHS of~\eqref{eq:lob_1} is further bounded by the following lemma:
\begin{lemma}\label{lem:lob_pen_tr_stab}
The penalty term is bounded from above as
\begin{align}
 &
  \sumT (\psi_t(q_{t+1}) - \psi_{t+1}(q_{t+1}))
  +
  \psi_{T+1}(u)
  -
  \psi_1(q_1)
 \nn
 &\quad
  \leq 
  \gamma \sumak \beta'_{T+1, a} 
  +
  4 m k^2 \gamma 
  \com 
\end{align}
and
the optimal value of ExO is bounded as
\begin{equation}
  \OPT_{q_t}(\beta_t)
  \leq
  2 m^2 k^2 \sumak \frac{\alpha_{ta}}{\beta'_{ta}}
  \per 
\end{equation}
\end{lemma}
The bound for the penalty term follows from $u \geq \ones/T$, which is standard in the literature.
The bound for $\OPT_{q_t}(\beta_t)$ is a direct application of Lemma~\ref{lem:stab_bound_pm_lb}.

Now, we are ready to prove Theorem~\ref{thm:lob-main-theorem}.
Here, we only sketch the proof and the complete proof can be found in Appendix~\ref{subsec:proof-lob-main-thm}.
\begin{proof}[Proof sketch of Theorem~~\ref{thm:lob-main-theorem}]
For notational simplicity, we may ignore terms whose dependence on $T$ is of order less than $\log T$ in this sketch.
We first bound $\sumak \frac{\alpha_{ta}}{\beta'_{ta}}$ in Lemma~\ref{lem:lob_pen_tr_stab} with the definitions of $\beta_t$ and $\alpha_t$.
Since $\alpha_0 \geq \frac{1}{\gamma^{3/2}} \geq \frac{1}{\gamma} \alpha_t$,
one can prove that
$
  \sumT \frac{\alpha_{ta}}{\beta'_{ta}} 
  \lesssim
  O\prx{\frac{\gamma}{c^2} \prx{ \beta'_{T+1, a} - \beta'_{1,a} } } 
.
$
Combining this inequality with Lemmas~\ref{lem:lob_1} and~\ref{lem:lob_pen_tr_stab} and $c = 2mk$ gives
$
  \Reg_T
  =
  O\prn[\big]{
  \E\brk{
    \gamma
    \sumT \sumak 
    \beta'_{T+1, a}
  }
  }
  .
$

We next bound $\beta'_{T+1,a}$ for suboptimal action $a \neq a^*$ and optimal actions $a^*$ separately.
From the definition of $\beta'_{ta}$ in~\eqref{eq:def_lr_local},
we have
$
  \beta'_{T+1,a}
  \leq 
  c \sqrt{ \alpha_0  + {\sum_{t=1}^{T} q_{ta} }/{\gamma}}
$
for all $a \neq a^*$.
For optimal action $a^*$,
using the inequality
$
  \min  
  \crl[\big]{
    z
    ,\,
    \frac{1 - z}{\gamma z}
  }
  \leq 
  \frac{2}{\sqrt{\gamma}} (1 - z)
$
that holds for any $z \in [0,1]$ due to $\sqrt{\gamma} \geq 2$,
we have
$
  \sumT \alpha_{ta^*}  
  \leq
  \frac{2}{\sqrt{\gamma}} \sumT (1 - q_{ta^*})
  =
  \frac{2}{\sqrt{\gamma}} \sumT \sum_{a \neq a^*} q_{ta}
  \per 
$
This implies that $\beta'_{T+1, a^*}$ is bounded as
$
  \beta'_{T+1, a^*}
  \leq
  c \sqrt{ \alpha_0 + \frac{2}{\gamma^{3/2}} \sumT \sum_{a \neq a^*} q_{ta} }
  \per 
$
Combining the upper bounds on $(\beta'_{ta})_{a \neq a^*}$ and $\beta'_{ta^*}$ implies that the regret is roughly bounded as 
$
  \Reg_T
  \leq   
  O\prn{
  \gamma 
  \sum_{a \neq a^*}
  \E\brk*{
    c \sqrt{
    \sumT \frac{q_{ta}}{\gamma}}
  }
  +
  \gamma c \sqrt{ 
  \E\brk*{
  \sum_{t, a\neq a^*}
  \frac{q_{ta}}{\gamma^{3/2}} }
  }
  }
  \leq    
  O\prn*{
  \gamma
  \sum_{a \neq a^*}
  c \sqrt{
  \frac{Q_a}{\gamma} }
  +
  \gamma c \sqrt{ 
  \sum_{a \neq a^*} \!\! \frac{Q_a}{{\gamma^{3/2}}} }
  }
\com 
$
where in the last inequality we let $Q_a = \E[\sumT q_{ta}]$.

For adversarial environments, using the Cauchy--Schwarz inequality and $\sumak q_{ta} = 1$ in the last inequality gives that the regret is roughly bounded as
$
  \Reg_T
  =
  O\prn[\big]{
  \gamma c \sqrt{  
  {Tk}/{\gamma}  }
  +
  \gamma c \sqrt{  
  {k T }/{\gamma^{3/2}} }
  }
  =
  O(
  m k^{3/2} \sqrt{T \log T }
  )
  \per 
$

For stochastic environments, since $p \in \calR(q)$, it holds that $\Reg_T \geq \frac{1}{2} \sum_{a \neq a^*} \Delta_a Q_a$.
Hence, 
$
  \Reg_T 
  = 
  2 \Reg_T - \Reg_T
  \leq 
  O\prn{
  \gamma 
  \sum_{a \neq a^*} 
  c \sqrt{\frac{Q_a}{\gamma} }
  +
  \gamma c \sqrt{ \sum_{a \neq a^*} \frac{Q_a}{\gamma^{\frac32}}}
  -  \sum_{a \neq a^*} \Delta_a Q_a
  }
  \leq    
  O\prn{
  \sum_{a \neq a^*}
  \prn[\big]{
    c \sqrt{ \gamma Q_a }
    -
    \Delta_a Q_a
  }
  +
  c \prn[\big]{ { \sqrt{\gamma}} \sum_{a \neq a^*}  Q_a }^{\frac12}
  - \Deltamin \!\!\sum_{a \neq a^*}  Q_a
  }
\per 
$
Taking the worst case with respect to $(Q_a)_{a \neq a^*}$ in the last inequality completes the proof for stochastic environments.
The proof for stochastic environments with adversarial corruptions mostly follows the same arguments.
\end{proof}

\section{Globally Observable Games}\label{sec:global}
This section presents an improved BOBW algorithm for globally observable games and shows its regret bounds.
Let $c_{\calG} = \max\{1, k \nrm{G^{\circ}}_{\infty} \}$ be the game-dependent constant.

We use FTRL with regularizer $\psi_t \colon \calP_k \to \R$ defined by
\begin{equation}\label{eq:defpsiglobal}
  \psi_t(q)
  = 
  \beta_t 
  \prx{ \psiLB(q) + \psiCLB(q) }
\end{equation}
for the log-barrier regularizer
$\psiLB(x) = - \sumak \log x_a$ and the complement log-barrier regularizer $\psiCLB(x) = - \sumak \log (1-x_a)$.
The learning rate $\beta_t \in \Rp$ in~\eqref{eq:defpsiglobal} is  
\begin{equation}
  \beta_t
  =
  c_1 \prx{k + \sum_{s=1}^{t-1} z_s }^{2/3}
  \!\mbox{with}\,\,
  z_t = \sumak \min \{ q_{ta} , 1 - q_{ta} \}
 \n
\end{equation}
for $c_1 > 0$ specified later.
The output of FTRL $q_t$ is transformed to action selection probability $p_t \in \calP_k$ by
\begin{equation}\label{eq:p_global}
  p_t = (1 - \gamma_t) q_t + \gamma_t \frac{\ones}{k}
  \quad \mbox{with} \quad
  \gamma_t
  =
  \frac{\sqrt{2} c_{\calG} z_t }{\sqrt{\beta_t}}
  \per 
\end{equation}
We use the loss estimator 
$
  \hat{y}_{t} = {G^\circ(A_t, \Phi_{A_t x_t})}/{p_{tA_t}}
$
for $G^\circ$ in~\eqref{eq:G_0}.
The proposed algorithm for globally observable games is summarized in Algorithm~\ref{alg:global} in Appendix~\ref{sec:proof-global}.

The above algorithm achieves the following bounds:
\begin{theorem}\label{thm:gob-main-theorem}
Consider any globally observable partial monitoring games.
In stochastic environments, the above algorithm (Algorithm~\ref{alg:global} in Appendix~\ref{sec:proof-global}) with $c_1 = \prn{{3 c_{\calG}}/{(k \log T) }}^{2/3}$ achieves
\begin{equation}
  \Reg_T 
  = 
  O \prx{
    \frac{c_{\calG}^2 k \log T}{\Delta^2}
    +
    c_{\calG}^{2/3} k^{4/3} \prx{\log T}^{1/3}
  }
  \per 
  \label{eq:regret_global_stoc}
\end{equation}
In adversarial environments, the same algorithm achieves
\begin{equation}
  \Reg_T 
  =
  O\prn*{
    \prn{c_{\calG}^2 k \log T}^{1/3} T^{2/3}
    +
    \prn{c_{\calG}^2 \log T}^{1/3} k
  }
  \per 
 \n 
\end{equation}
Further, in stochastic environments with adversarial corruptions, the same algorithm achieves $\Reg_T = O(\calR^{\mathrm{glo}} + \prn{C^2 \calR^{\mathrm{glo}}}^{1/3})$ with $\calR^{\mathrm{glo}}$ the RHS of~\eqref{eq:regret_global_stoc}.
\end{theorem}
The proof of Theorem~\ref{thm:gob-main-theorem} can be found in Appendix~\ref{sec:proof-global}.
This is the first BOBW bound for globally observable games with an $O(\log T)$ stochastic bound.
The above result improves the existing BOBW bounds for PM in stochastic environments, where the bound depends on $\log T$ rather than~$\log^2 T$, although its dependence on $k$ is $k$-times worse.
The bound for adversarial environments is~$\Theta(\sqrt{\log(kT)})$-times smaller than the bound in~\citet{tsuchiya23best}.

The aim of this section is to demonstrate the usefulness of hybrid regularizers by establishing the first $O(\log T)$ stochastic bounds.
The choice of the regularizer and analysis largely follow~\citet{ito22revisiting} and there is no significant novelty in this section from the technical viewpoint.
As a small contribution, we revised the proof of their paper for dealing with the hybrid regularizer and provided the correct proof.\footnote{In~\citet[Lemma 4.9, Proposition 4.10]{ito22revisiting}, a case distinction based on the output of FTRL $q$ needs to be discussed.
This paper addresses and corrects this, as detailed in Appendix~\ref{sec:proof-global}.}
Note also that one can introduce ExO by using a similar analysis for the hybrid regularizer introduced in Section~\ref{sec:exo} in globally observable games, 
although we do not provide the details as it does not improve the worst-case bound.

\section{Concluding Remarks}\label{sec:conclusion}
In this paper, we proposed the first BOBW algorithms for PM that achieve $O(\log T)$ regret bounds in stochastic environments for both locally observable and globally observable games. 
This was accomplished through the use of hybrid regularizers known to be useful in simpler online learning problems,
and the new analysis to combine these regularizers with PM problems, in which we need to deal with limited feedback.
The most important contribution of this paper is the development of the analysis to upper-bound the optimal value of the ExO objective induced by the hybrid regularizer over restricted feasible region $\calR(q)$. 
This analysis enabled us to significantly improve the existing regret upper bounds in the stochastic environments.

Our regret upper bounds match the known lower bounds in terms of $T$ both in the worst-case sense~\cite{Bartok11minimax,lattimore19cleaning} and in the gap-dependent sense~\cite{Komiyama15PMDEMD}.
However, little else is understood about the dependencies on the variables $k$, $d$, and $m$, as mentioned in \citet[Section 37.9]{lattimore2020book}.
Investigating these dependencies is an important future work.

\section*{Acknowledgements}
TT was supported by JST, ACT-X Grant Number JPMJAX210E, Japan. JH was supported by JSPS, KAKENHI Grant Number JP21K11747, Japan.

\section*{Impact Statement}
This work is primarily theoretical and is not expected to have any direct negative ethical or societal consequences. 
The result could be advantageous for those studying theoretical elements of online learning with limited feedback, such as online learning with feedback graphs and partial monitoring. 
Additionally, the improved regret bounds may contribute to the frequent use of partial monitoring algorithms in real-world applications such as recommendation systems and dynamic pricing.

\bibliography{ref.bib}
\bibliographystyle{icml2024}

\newpage
\appendix
\onecolumn

\section{Notation}\label{sec:notation}

Table~\ref{table:notation} summarizes the notation used in this paper.

\begin{table}[h]
  \caption{
  Notation
  }
  \label{table:notation}
  \centering
  \small
  \begin{tabular}{ll}
    \toprule
    Symbol & Meaning  \\
    \midrule
    $\calP_k$ & $(k-1)$-dimensional probability simplex \\
    $D_{\psi}(x, y)$ & Bregman divergence induced by $\psi$ from $y$ to $x$ \\
    $\psi^*$ & convex conjugate of $\psi$ \\
    \midrule
    $T \in \N$ & time horizon \\
    $k \in \N$ & number of actions \\
    $d \in \N$ & number of outcomes  \\ 
    $\lossmat \in [0,1]^{k \times d}$ & loss matrix  \\ 
    $\Sigma$ & set of feedback symbols  \\ 
    $\fbmat \in \Sigma^{k \times d}$ & feedback matrix  \\ 
    $m \in \N$ & maximum number of distinct symbols in a single row of $\Phi$ \\ 
    $a^* \in \Pi$ & (unique) optimal action \\
    $A_t \in [k]$ & action chosen by learner at round $t$  \\
    $x_t \in [d]$ & outcome chosen by adversary at round $t$  \\
    \midrule
    $\mathrm{path}_{\mathscr{T}}$ & path over in-tree $\mathscr{T}$ \\
    $\calH$  & the set of all functions from $[k] \times \Sigma$ to $\R^k$  \\
    $\calH^{\circ}$  & the set of $G \colon [k] \times [d] \to \R^k$ satisfying~\eqref{eq:Gdiff_Ldiff} \\
    $G^{\circ} \in \calH^{\circ}$ & an example of $G$ in $\calH^{\circ}$ defined in \eqref{eq:G_0} \\

    $c_{\calG} > 0$ & game-dependent constant defined in Section~\ref{sec:global} \\
    $\Delta_a \in [0,1]$ & suboptimality gap of action $a$ \\  
    \midrule
    $q_t \in \calP_k$ & output of FTRL at round $t$   \\ 
    $p_t \in \calP_k$ & action selection probability at round $t$   \\ 
    $\psi_t \colon \calP_k \to \R$ & regularizer of FTRL at round $t$  \\ 
    $\beta_t > 0$  & learning rate of FTRL at round $t$  \\ 
    $\psiNS \colon \R_+^k \to \R $ & $\sumk x_i \log x_i $, negative Shannon entropy  \\ 
    $\psiCNS \colon (-\infty, 1)^k \to \R $ & $\sumk (1-x_i) \log (1-x_i) $, complement negative Shannon entropy  \\ 
    $\psiLB \colon \R_+^k \to \R $ & $\sumk \log(1/x_i) $, log-barrier \\ 
    $\psiCLB \colon (-\infty, 1)^k \to \R $ & $\sumk \log(1/(1-x_i)) $, complement log-barrier \\ 
    \midrule 
    $\stab^\psi_q \colon \R^k \to \R$ & stability function induced by function $\psi \colon \R^k \to \R$ in~\eqref{eq:def_calS}  \\
    $\calP'(q) \subset \calP_k$ & feasible set used in existing ExO defined in~\eqref{eq:exo-bobwpm}  \\
    $\calR(q) \subset \calP_k$ & feasible set used in ExO defined in~\eqref{eq:half-truncated-simplex}  \\
    $\objexo(\,\cdot\,; q, (\beta, \phi))$ & objective function in ExO with regularizer $\psi(q) = \sum_{a=1}^k \beta_a \phi(q_a)$ in~\eqref{eq:exo-obj}  \\
    $\OPT_q(\beta)$ & optimal value of ExO objective $\objexo$ minimized over $\calR(q)$ \\ 
    \bottomrule
  \end{tabular}
\end{table}
\section{Missing Related Work}\label{sec:additional-related-work}

\paragraph{Exploration by Optimization}
Exploration by optimization (ExO), an important technique for achieving a good regret upper bound in locally observable games, was invented by~\citet{lattimore20exploration} and further extended by~\citet{lattimore21mirror}.
Very recent studies have revealed that ExO is closely related to the notion of \emph{(convexified) decision-estimation coefficient} and the optimal regret upper bound in adversarial environments is characterized by the ExO framework~\citep{foster22complexity}.
In the context of BOBW algorithms, ExO was very recently investigated in PM~\citep{tsuchiya23best, tsuchiya23stability} and linear bandits~\citep{ito23exploration}.

\paragraph{Best-of-both-worlds Guarantees}
The investigation of the BOBW algorithm was first initiated by \citet{bubeck2012best} in multi-armed bandits, and after that, it has been shown that the BOBW guarantees are possible in a wide range of online decision-making problems.
Many algorithms have been developed for special cases of PM such as predication with expert advice~\citep{gaillard2014second, luo2015achieving}, multi-armed bandits~\citep{zimmert19optimal, zimmert2021tsallis, ito21parameter}, and online learning with feedback graphs~\citep{rouyer2022near, ito2022nearly}.
Most BOBW algorithms are based on FTRL (or online mirror descent) because the self-bounding technique  
invented in~\citet{gaillard2014second} and used in many subsequent studies,~\eg\citet{wei2018more, zimmert2021tsallis}, is a strong tool to derive (poly-)logarithmic regret bounds in stochastic environments.

It is worth noting that the very recent work by~\citet{dann23blackbox} has devised a general framework that transforms a bandit algorithm for adversarial environments into a BOBW algorithm in a blackbox manner. 
However, it is highly nontrivial whether their algorithm can be applied to PM, and a detailed discussion can be found in Section~\ref{sec:local}.

The use of FTRL with the hybrid regularizer consisting of the log-barrier and complement Shannon entropy is not new: This regularizer has been used to obtain BOBW guarantees in multi-armed bandits and combinatorial semi-bandits~\citep{ito2021hybrid, ito22adversarially, tsuchiya23further}.
The idea of using complement Shannon entropy was originally introduced by~\citet{zimmert2019beating} in the context of combinatorial semi-bandits.
Note, however, that PM is much more complex than these problems, and a non-trivial analysis is required when combining this regularizer with the ExO approach as detailed in Section~\ref{sec:exo}.

\section{Stochastic Environment with Adversarial Corruptions}\label{sec:corrupted_env}
Here, we introduce a stochastic environments with adversarial corruptions, considered originally in multi-armed bandits~\citep{lykouris2018stochastic} and later investigated also in PM~\citep{tsuchiya23best, tsuchiya23stability}.
This environment is an intermediate environment between the stochastic environments and adversarial environments.

In this environment, at each time step $t \in [T]$,
a temporary outcome $x'_t \in [d]$, which cannot be observed by the learner, is sampled from an unknown distribution $\nu^*$ in an i.i.d.~manner.
The adversary then corrupts $x'_t$ to $x_t$ without knowing $A_t$, and based on this outcome the loss and feedback symbol observed by the learner is determined.
In this environment, the subotimality gap is defined in terms of $(x'_t)$ as follows: 
 $\Delta_a = \E_{x'_t \sim \nu^*} [ \calL_{a x'_t} - \calL_{a^* x'_t}  ]$ for $a \in [k]$.

The corruption level $C \geq 0$ in this environment is defined by
$
  C 
  = 
  \E\brk{\sum_{t=1}^T \nrm{\calL e_{x_t} - \calL e_{x'_t}}_\infty}
  \per 
$
Note that the stochastic environment with adversarial corruptions with $C = 0$ (resp.~$C = 2T$) corresponds to the stochastic (resp.~adversarial) environment.
Note that the proposed algorithms will work without the knowledge of $C$. 

In addition to the stochastic environment with adversarial corruptions, there are several intermediate environments, a stochastically constrained adversarial environment, and an adversarial environment with a self-bounding constraint that includes all of these environments as special cases. 
Detailed definitions can be found in~\citet{tsuchiya23best}.
The algorithms and bounds in this paper are directly applicable to these general environments as well.
For the sake of brevity, this paper focuses only on the stochastic environments with adversarial corruptions.

Under this assumption, the regret in the stochastic environment with adversarial corruptions is bounded from below as follows:
\begin{lemma}\label{lem:lb_corrupt}
Consider any stochastic environment with adversarial corruptions.
Then, the regret is bounded from below as
\begin{equation}
  \Reg_T 
  \geq
  \E\brk*{ \sumT \sum_{a \neq a^*} \Delta_a \, \pta } - 2 \, C
  \per 
  \n
\end{equation}
\end{lemma}
This lower bound will be used to derive the logarithmic regret bound for stochastic environments.
The proof can be found in Appendix A of~\citet{tsuchiya23best},
and we include the proof for completeness.
\begin{proof}[Proof of Lemma~\ref{lem:lb_corrupt}]
  Recall that in the stochastic environment with adversarial corruptions,
  $x'_1, x'_2, \dots$ are sampled in an i.i.d.~manner and the suboptimality gap is defined by
   $\Delta_a = \E_{x'_t \sim \nu^*} [ \calL_{a x'_t} - \calL_{a^* x'_t}  ]$ for $a \in [k]$.
   Hence, the regret is bounded from below as
  \begin{align}
    \Reg_T
    &=
    \Expect{ \sumT \prx{\calL_{A_t x_t} - \calL_{a^* x_t}} }  
    \nn
    &=
    \Expect{ \sumT \prx{\calL_{A_t x'_t} - \calL_{a^* x'_t}} }  
    +
    \Expect{ \sumT \prx{\calL_{A_t x_t} - \calL_{A_t x'_t}} }  
    +
    \Expect{ \sumT \prx{\calL_{a^* x'_t} - \calL_{a^* x_t}} }  
    \nn
    &\geq 
    \Expect{ \sumT \sum_{a \neq a^*} \Delta_a \, \pta}
    - 
    2 \, C
    \com 
    \n
  \end{align}
  which completes the proof of Lemma~\ref{lem:lb_corrupt}.
  \end{proof}

\section{Common Analysis}\label{sec:proof-common}

\subsection{Standard Analysis of FTRL}
\begin{proof}[Proof of Lemma~\ref{lem:ftrl}]
Let $F_t(q) = \sum_{s=1}^{t-1} \innerprod{\hat{y}_s}{q} + \psi_t(q)$ be the objective function of FTRL at time $t$.
Then it holds that 
\begin{align}
  \sumT \innerprod{\hat{y}_t}{q_t - u}
  \leq 
  \psi_{T+1}(u) - \psi_1(x_1)
  +
  \sumT \prx{ F_t(q_t) - F_{t+1}(q_{t+1}) + \innerprod{\hat{y}_t}{q_t}}
  \per
  \label{eq:p_lem_ftrl_1}
\end{align}
This inequality holds since
\begin{align}
  - \sumT \innerprod{\hat{y}_t}{u}
  &= 
  \psi_{T+1}(u) - F_{T+1}(u)
  +
  \sumT \prx{ F_t(q_t) - F_{t+1}(q_{t+1})}
  - F_1(q_1) + F_{T+1}(q_{T+1})
  \nn
  &\leq
  \psi_{T+1}(u) - \psi_1(q_1)
  +
  \sumT \prx{ F_t(q_t) - F_{t+1}(q_{t+1})}
  \com 
  \n
\end{align}
where the equality holds by the definition of $F_t$.
Now it remains to bound $F_t(q_t) - F_{t+1}(q_{t+1}) + \innerprod{\hat{y}_t}{q_t}$ in~\eqref{eq:p_lem_ftrl_1}.
This is bounded as
\begin{align}
  &
  F_t(q_t) - F_{t+1}(q_{t+1}) + \innerprod{\hat{y}_t}{q_t}
  \nn
  &=
  F_t(q_t) - F_{t}(q_{t+1}) 
  +
  \psi_{t}(q_{t+1}) - \psi_{t+1}(q_{t+1})
  + 
  \innerprod{\hat{y}_t}{q_t - q_{t+1}}  
  \nn
  &\leq
  - D_{F_t} (q_{t+1}, q_t) 
  +
  \psi_{t}(q_{t+1}) - \psi_{t+1}(q_{t+1})
  + 
  \innerprod{\hat{y}_t}{q_t - q_{t+1}}
  \com
  \n
\end{align}
where the inequality follows from the first-order optimality condition.
Since $D_{F_t}(q_{t+1}, q_t) = D_{\psi_t}(q_{t+1}, q_t)$ the proof is completed.
\end{proof}

\subsection{Basic Facts to Bound Stability Terms}\label{sec:facts_for_bound_stab}
This section presents the basic propositions to bound stability terms, which are quite standard in the literature.
The stability terms are bounded in terms of the following functions:
\begin{align}
  \xi(x) &\coloneqq \exp(-x) + x - 1 
  \leq 
  \begin{cases}
    \frac12 x^2 &\, \mbox{for}\;  x \geq 0  \\
    x^2         &\, \mbox{for}\;  x \geq -1 \com 
  \end{cases}
  \label{eq:bound_xi}
  \\
  \zeta(x) &\coloneqq x - \log(1 + x) \leq x^2 \quad \mbox{for} \;  x \geq -\frac{1}{2}
  \label{eq:bound_zeta} 
  \per 
\end{align}

Recall that 
$\phiCNS(x) = (1-x) \log (1-x)$,
$\phiLB(x) = - \log x$, and $\phiCLB(x) = - \log (1-x)$
are the components of the complement negative Shannon entropy, log-barrier, and complement log-barrier, respectively.

\begin{proposition}[stability with complement negative Shannon entropy]\label{prop:stab_cns}
For $x \in (0,1)$,
\begin{align}
  \max_{y \in (-\infty, 1)} \brx{ a(x - y) - D_{\phiCNS}(y,x) } 
  &= (1-x) \xi(-a) 
  \quad \mbox{for} \quad 
  a \in \R
  \com 
  \n
\end{align}
where $\phiCNS(x) = (1-x)\log(1-x)$.
The last quantity is further bounded by $(1-x)a^2$ if $a \leq 1$.
\end{proposition}
\begin{proof}
Since $ a(x - y) - D_{\phiCNS}(y,x) $ is concave with respect to $y \in (-\infty, 1)$, the maximizer $y^*$ satisfies
$
  - a - \nabla \phiCNS(y^*) + \nabla \phiCNS(x) = 0  
  \com 
$
which is equivalent to  (since $\nabla \psiCNS(x) = - \log (1-x) - 1$)
\begin{align}
  - a + \log(1-y^*) - \log (1-x) = 0
  \per 
  \n
\end{align}
Hence $1 -y^* = \e^a (1-x)$.
Hence
\begin{align}
  &
  a(x - y^*) - D_{\phiCNS}(y^*,x)
  = D_{\phiCNS}(x, y^*)
  \nn 
  &= 
  (1-x) \log (1 - x) - (1-y^*) \log(1 - y^*) + (\log(1-y^*) + 1)(x - y^*)
  \nn
  &= 
  (1-x) \log (1 - x) - \log(1 - y^*) + x (\log(1-y^*) + 1) - y^*
  \nn
  &=
  - a (1-x) + x - y^*
  = (1-x) (\e^a - a - 1) = (1-x) \xi(-a)
  \per 
  \n
\end{align}
If $-a \geq -1$ holds, then from~\eqref{eq:bound_xi} the last quantity is further bounded by $(1-x) \xi(-a) \leq (1-x) a^2$.
\end{proof}

\begin{proposition}[stability with log-barrier]\label{prop:stab_lb}
For $x \in (0,1)$,
\begin{align}
  \max_{y \in \R_+} \brx{ a(x - y) - D_{\phiLB}(y,x) } &= \zeta(ax)
  \quad 
  \mbox{for}
  \quad 
  a \geq - \frac1x
  \com 
  \n
\end{align}
where $\phiLB(x) = - \log x$.
The last quantity is further bounded by $\zeta(ax) \leq x^2 a^2$ if $a \geq -\frac{1}{2x}$.
\end{proposition}
\begin{proof}
Since $ a(x - y) - D_{\phiLB}(y,x) $ is concave with respect to $y \in \R_+$, the maximizer $y^*$ satisfies
$
  - a - \nabla \phiLB(y^*) + \nabla \phiLB(x) = 0  
  \com 
$
which is equivalent to
\begin{align}
  - a + \frac{1}{y^*} - \frac{1}{x} = 0
  \per 
  \n
\end{align}
Note that $y^*$ satisfied $y^* > 0$ holds if $a + \frac{1}{x} > 0$, which is equivalent to $a > - \frac{1}{x}$. 
Hence
\begin{align}
  &
  a(x - y^*) - D_{\phiLB}(y^*,x)
  = D_{\phiLB}(x, y^*)
  \nn 
  &= 
  - \log (x) + \log(y^*) + \frac{x - y^*}{y^*}
  =
  - \log\prx{\frac{x}{y^*}} + \frac{x}{y^*} - 1 
  \nn
  &=
  - \log (1 + ax) + ax
  =
  \zeta(ax)
  \nn  
  &\leq 
  x^2 a^2
  \com 
  \n
\end{align}
where the inequality holds by
$- \log (1+z) + z \leq z^2$ for $z \geq -1/2$ (in \eqref{eq:bound_zeta}) combined with $a \geq -\frac{1}{2x}$.
\end{proof}

\begin{proposition}[stability with complement log-barrier]\label{prop:stab_clb}
For $x \in (0,1)$,
\begin{align}
  \max_{y \in (-\infty, 1)} \brx{ a(x - y) - D_{\phiCLB}(y,x) } &= \zeta(-a(1-x))
  \quad 
  \mbox{for}
  \quad a < \frac{1}{1-x}
  \com 
  \n
\end{align}
where $\phiCLB(x) = - \log(1-x)$.
The last quantity is further bounded by $\xi(-a(1-x)) \leq (1-x)^2 a^2$ if $a \leq \frac{1}{2(1-x)}$.
\end{proposition}
\begin{proof}
Since $ a(x - y) - D_{\phiCLB}(y,x) $ is concave with respect to $y \in (-\infty, 1)$, the maximizer $y^*$ satisfies
$
  - a - \nabla \phiCLB(y^*) + \nabla \phiCLB(x) = 0  
  \com 
$
which is equivalent to 
\begin{align}
  - a - \frac{1}{1-y^*} + \frac{1}{1-x} = 0
  \per 
  \n
\end{align}
Note that $ y^* < 1$ holds if $a < \frac{1}{1-x}$.
Hence
\begin{align}
  &
  a(x - y^*) - D_{\phiCLB}(y^*,x)
  = D_{\phiCLB}(x, y^*)
  \nn 
  &= 
  - \log (1 - x) + \log(1 - y^*) - \frac{x - y^*}{1 - y^*}
  =
  - \log\prx{\frac{1-x}{1-y^*}} + \frac{1-x}{1-y^*} - 1 
  \nn
  &=
  - \log (1 - a(1-x)) - a(1-x) 
  =
  \zeta(-a(1-x))
  \nn 
  &\leq 
  (1-x)^2 a^2
  \com 
  \n
\end{align}
where the inequality holds by
$- \log (1+z) + z \leq z^2$ for $z \geq -1/2$ (in \eqref{eq:bound_zeta}) combined with $a \leq \frac{1}{2(1-x)}$.
\end{proof}

\section{Deferred Proofs for Exploration by Optimization (Lemma~\ref{lem:stab_bound_pm_lb}, Section~\ref{sec:exo})}\label{sec:proof-stab-wto}

Before going to the proof of Lemma~\ref{lem:stab_bound_pm_lb}, we introduce the water transfer operator.

\begin{lemma}[\citealp{lattimore2019information}]\label{lem:wto}
Consider any non-degenerate and locally observable partial monitoring games and let $\nu \in \calP_d$. 
Then there exists a function $W_\nu \colon \calP_k \to \calP_k$ such that the following three statement hold for all $q \in \calP_k$:
\textup{(a)} $\left(W_\nu(q) - q\right)^\top \calL \nu \leq 0$;\
\textup{(b)} $W_\nu(q)_a \geq q_a/k$ for all $a \in [k]$;\ and
\textup{(c)} there exists an in-tree $\mathscr{T} \subset \calE$ over $[k]$ such that $W_\nu(q)_a \leq W_\nu(q)_b$ for all $(a, b) \in \mathscr{T}$.
\end{lemma}

Using this water transfer operator, we prove Lemma~\ref{lem:stab_bound_pm_lb}.
\begin{proof}[Proof of Lemma~\ref{lem:stab_bound_pm_lb}]
Recall that
$
  \calH^{\circ}
  =
  \big\{
    G \colon
    (e_b - e_c)^\top \sumak G(a, \Phi_{ax})
    = 
    \calL_{bx} - \calL_{cx} \mbox{ for all } b, c \in \Pi \mbox{ and } x \in [d]
  \big\}
  \per 
$
Recall also that
$
  \stab_q^\phi(z) 
  = 
  D_{\phi^*}  (\nabla \phi(q) - z, \nabla \phi(q))
$
and $(p, g) \mapsto p \stab_q^\phi(g/p)$ is convex if $p > 0$.

\paragraph{Step 1. Applying Sion's minimax theorem}
Take any $q \in \calP_k$.
By Sion's minimax theorem, 
\begin{align}
  \OPT_q(\beta)
  &\leq 
  \min_{G \in \calH^{\circ},\, p \in \calR(q)} 
  \max_{\nu \in \calP_d} 
  \left[
  (p - q)^\top \calL \nu
  + 
  \sum_{x=1}^d 
  \nu_x 
  \sum_{b=1}^k \beta_b
  \sumak 
  p_a 
  \stab_{q_b}^\phi\prx{ \frac{ G(a, \Phi_{ax})_b }{ \beta_b \, p_a } }
  \right]
  \nn 
  &\leq 
  \max_{\nu \in \calP_d} 
  \min_{G \in \calH^{\circ},\, p \in \calR(q)} 
  \left[
  (p - q)^\top \calL \nu 
  + 
  \sum_{x=1}^d \nu_x 
  \sum_{b=1}^k \beta_b
  \sumak 
  p_a 
  \stab_{q_b}^\phi\prx{ \frac{{G}(a, \Phi_{ax})_b  }{ \beta_b \, p_a} }
  \right] 
  \per 
  \label{eq:maxmin_obj}
\end{align}

Take any $\nu \in \calP_d$ and let $\mathscr{T}$ be the in-tree over $[k]$ induced by water transfer operator $W_\nu \colon \calP_k \to \calP_k$ by Part (c) of Lemma~\ref{lem:wto}.
Then we let $ u = W_{\nu}(q) $.
Using this $u$, we define $p \in \calR(q)$ by 
\begin{align}
p = (1 - \lambda) u + \lambda q
\quad
\mbox{with}
\quad 
\lambda = 1/2  
\per 
\n
\end{align}
By the definition of $p$, it holds that
$p_a \geq u_a/2 = W_\nu(q)_a /2$
and 
$p_a \geq q_a/ 2$,
and it indeed holds that $p \in \calR(q)$.
We can take $G \in \calH^{\circ}$ such that 
$
G(a, \sigma)_b = \sum_{e \in \mathrm{path}_{\mathscr{T}}(b)} w_e(a, \sigma)
$
with $w_e$ satisfying $\nrm{w_e}_\infty \leq m/2$ since $\calG$ is non-degenerate~\citep[Lemma 20]{lattimore20exploration}.
Since paths in $\mathscr{T}$ have length at most $k$, we have $\nrm{G}_\infty \leq km/2$.

\paragraph{Step 2. Preliminaries for bounding the stability term}
In the following, we first consider the stability term, which corresponds to the second term in~\eqref{eq:maxmin_obj}.
Fix $b \in [k]$ and let $z_b = G(a, \Phi_{ax}) / (\beta_b \, p_{a}) \in \R$ for notational simplicity (ignoring index $a$).
Taking $p' \in \R$ such that $\nabla \phi(p') = \nabla\phi(q_b) - z_b$,
we have 
\begin{align}
  \stab^\phi_{q_b}(z_b)
  &=
  D_{\phi^*}  (\nabla \phi(q_b) - z_b, \nabla \phi(q_b))
  \nn 
  &=
  \phi^* (\nabla \phi(q_b) - z_b) 
  - 
  \phi^*(\nabla \phi(q_b))
  -
  \nabla \phi^* ( \nabla \phi(q_b) ) \cdot\prn{-z_b}
  \nn 
  &=
  - D_{\phi^*}(\nabla \phi(q_b), \nabla \phi(p'))
  - 
  {\nabla \phi^*( \nabla (p')) } \cdot \prn{\nabla \phi(q_b) - \nabla \phi(p')}
  +
  {q_b}{z_b}
  \tag{by $\nabla \phi^* = (\nabla \phi)^{-1}$}
  \nn
  &=
  - D_{\phi}(p', q_b)
  - 
  {p'}{z_b}
  +
  {q_b}{z_b}
  \tag{by $D_{\phi}(x, y) = D_{\phi^*}(\nabla \phi(y), \nabla \phi(y))$}
  \nn
  &=
  \prn{q_b - p'}{z_b} - D_{\phi}(p', q_b)
  \per 
  \n
\end{align}
Hence, by the linearity of the Bregman divergence and $\psi(q) = \sum_{b=1}^k \beta_b \phi(q_b) = \sum_{b=1}^k \beta_b (\phiLB (q_b) + \gamma \phiCNS (q_b))$ in~\eqref{eq:defpsiexo},
\begin{align}\label{eq:stab_decompose}
  \stab^\phi_{q_b}(z)
  &= 
    (q_b - p'_b) z_b
    -
    \beta_b
    \big(
      D_{\phiLB} (p_b', q_b)
      +
      \gamma
      D_{\phiCNS} (p_b', q_b)
    \big)
  \nn
  &\leq 
  \min 
  \brx{
    (q_b - p'_b) z_b
    -
    \beta_b
    D_{\phiLB} (p_b', q_b)
    ,\,
    (q_b - p'_b) z_b
    -
    \gamma
    \beta_b
    D_{\phiCNS} (p_b', q_b)
  }
  \per 
\end{align}

We next check the condition to bound the stability term.
In particular, we will prove for any $a, b \in [k]$ it holds that
\begin{align}
  \abs[\bigg]{
    \frac{q_b G(a, \Phi_{ax})_b}{\beta_b p_a}
  }
  \leq 
  \frac{2 m k}{\beta_b}
  \leq 
  \frac12
  \com 
  \label{eq:zeta_bound_condition}
\end{align}
which is the condition for bounding the stability term induced by the log-barrier regularizer.
This inequality holds since 
\allowdisplaybreaks
\begin{align} 
  \abs[\bigg]{
    \frac{q_b G(a, \Phi_{ax})_b}{\beta_b p_a}
  }
  &\leq 
  \frac{2 q_b}{\beta_b u_a}
  \abs{
    G(a, \Phi_{ax})_b
  }
  \tag{by $p_a \geq u_a / 2$}
  \nn
  &=
  \frac{2 q_b}{\beta_b u_a}
  \abs[\Bigg]{
    \sum_{e \in \mathrm{path}_\mathscr{T}(b)} w_e(a, \Phi_{ax})
  }
  \nn
  &\leq 
  \frac{m q_b}{\beta_b u_a}
  \abs[\Bigg]{
    \sum_{e \in \mathrm{path}_\mathscr{T}(b)} 
    \ind{a \in \calN_e}
  }
  \tag{by $\nrm{w_e}_\infty \leq m /2$}
  \nn 
  &\leq 
  \frac{2 m q_b}{\beta_b u_a}
  \ind{ a \in \cup_{e \in \mathrm{path}_\mathscr{T}(b)} \calN_e }
  \tag{since any action is in $\calN_e$ at most two edges in $\mathrm{path}_\mathscr{T}(b)$}
  \nn 
  &\leq 
  \frac{2 m q_b}{\beta_b u_b}
  \ind{ a \in \cup_{e \in \mathrm{path}_\mathscr{T}(b)} \calN_e }
  \tag{by $u_a \geq u_b$ for $a \in \mathrm{path}_\mathscr{T}(b)$ since Part (c) of Lemma~\ref{lem:wto}}
  \nn 
  &\leq 
  \frac{2 m k}{\beta_b}
  \ind{ a \in \cup_{e \in \mathrm{path}_\mathscr{T}(b)} \calN_e }
  \tag{by Part (b) of Lemma~\ref{lem:wto}}
  \nn 
  &\leq 
  \frac12
  \com 
  \n
\end{align}
where the last inequality follows by $\beta_b \geq 4 m k$.

\paragraph{Step 3. Bounding the stability term}
Hence, taking the worst-case with respect to $p'$ in~\eqref{eq:stab_decompose} with inequality~\eqref{eq:zeta_bound_condition}, 
we can further bound the part of the second term in~\eqref{eq:maxmin_obj} as
\begin{align}
  \sum_{b=1}^k \beta_b
  \sumak 
  p_a 
  \stab_{q_b}^\phi \prx{ \frac{  {G}(a, \Phi_{ax})_b  }{ \beta_b \, p_a} }
  \leq 
  \sumak p_a 
  \sum_{b=1}^k
  \min  
  \brx{
    \beta_b\,
    \zeta \prx{
      \frac{q_b G(a, \Phi_{ax})_b}{\beta_b \, p_a}
    }
    ,\,
    \gamma \beta_b
    (1 - q_b)
    \,
    \xi \prx{
      -
      \frac{G(a, \Phi_{ax})_b}{\gamma \beta_b \, p_a}
    }
  }
  \com 
  \label{eq:stab_1}
\end{align}
where this holds from Proposition~\ref{prop:stab_lb} with
$\frac{z_b}{\beta_b} = \frac{G(a, \Phi_{ax})_b}{\beta_b p_a} \geq - 1 / q_b$ for all $b \in [k]$ that holds by~\eqref{eq:zeta_bound_condition}, and Proposition~\ref{prop:stab_cns}.

In the following we consider the inside of the summations in~\eqref{eq:stab_1}.

\noindent
\textbf{Case $\gamma q_b \leq 1$:}
First consider the case of $\gamma q_b \leq 1$.
Then it holds that 
$
\frac{1 - q_b}{\gamma q_b^2}
\geq 
\frac{1 - 1/\gamma}{\gamma (1/\gamma)^2}
=
\gamma - 1
\geq 
1
$
since $\gamma \geq 2$.
Using this and ~\eqref{eq:bound_zeta} with~\eqref{eq:zeta_bound_condition}, 
we can bound the part of~\eqref{eq:stab_1} by
\begin{align}
  \zeta\prx{
    \frac{q_b G(a, \Phi_{ax})_b}{\beta_b \, p_a}
  }
  &\leq 
  \prx{
    \frac{q_b G(a, \Phi_{ax})_b}{\beta_b \, p_a}
  }^2
  \label{eq:stab_2} 
  \\
  &\leq 
  \frac{1 - q_b}{\gamma  q_b^2}
  \prx{
    \frac{q_b G(a, \Phi_{ax})_b}{\beta_b \, p_a}
  }^2
  =
  \gamma (1 - q_b)
  \prx{
    \frac{G(a, \Phi_{ax})_b}{\gamma \beta_b \, p_a}
  }^2
  \label{eq:stab_3}
  \per
\end{align}

\noindent
\textbf{Case $\gamma q_b > 1$:}
Next we consider the case of $\gamma q_b > 1$.
Now we check the stability condition for the complement negative Shannon entropy.
Using~\eqref{eq:zeta_bound_condition} and $q_b \geq 1/\gamma$,
\begin{align}
  \abs[\bigg]{
    \frac{G(a, \Phi_{ax})}{\gamma \beta_b p_a}
  }
  =
  \frac{1}{\gamma q_b}
  \abs[\bigg]{
    \frac{q_b G(a, \Phi_{ax})}{\beta_b p_a}
  }
  \leq 
  \frac{2 m k}{\gamma \beta_b q_b}
  \leq 
  \frac{2 m k \gamma}{\gamma \beta_b}
  \leq 
  1
  \com 
  \label{eq:xi_bound_condition}
\end{align}
where the last inequality holds by $\beta_b \geq 4 m k$.
Hence, using~\eqref{eq:bound_zeta} with \eqref{eq:zeta_bound_condition} and \eqref{eq:bound_xi} with \eqref{eq:xi_bound_condition}, we can bound \eqref{eq:stab_1} by
\begin{align}
  \sumak p_a 
  \sum_{b=1}^k
  \beta_b
  \min  
  \brx{
    \prx{
      \frac{q_b G(a, \Phi_{ax})_b}{\beta_b \, p_a}
    }^2
    ,\,
    \gamma (1 - q_b)
    \,
    \prx{
      \frac{G(a, \Phi_{ax})_b}{\gamma \beta_b \, p_a}
    }^2
  }
  \per 
  \label{eq:stab_3_prime}
\end{align}

Summing up the arguments so far in~\eqref{eq:stab_2},~\eqref{eq:stab_3}, and~\eqref{eq:stab_3_prime},
we can bound~\eqref{eq:stab_1} by
\begin{align}
  &
  \sumak 
  p_a 
  \sum_{b=1}^k
  \beta_b
  \min  
  \brx{
    \prx{
      \frac{q_b G(a, \Phi_{ax})_b}{\beta_b \, p_a}
    }^2
    ,\,
    \gamma (1 - q_b)
    \,
    \prx{
      \frac{G(a, \Phi_{ax})_b}{\gamma \beta_b \, p_a}
    }^2
  }
  \nn
  &=
  \sumak
  \sum_{b=1}^k
  \frac{1}{\beta_b}
  \min  
  \brx{
    \frac{q_b^2}{p_a} 
    ,\,
    \frac{1 - q_b}{\gamma p_a}
  }
  G(a, \Phi_{ax})_b^2
  \per 
  \label{eq:stab_4}
\end{align}

\paragraph{Step 4. Further bounding the stability term based on the property of water transfer operator}

Using $p_a \geq u_a / 2$, $\nrm{w_e}_\infty \leq m / 2$, and Part (b) of Lemma~\ref{lem:wto},
and recalling that 
$
G(a, \sigma)_b = \sum_{e \in \mathrm{path}_{\mathscr{T}}(b)} w_e(a, \sigma) ,
$
\eqref{eq:stab_4} is bounded as
\allowdisplaybreaks
\begin{align} 
  &
  \sumak
  \sum_{b=1}^k
  \frac{1}{\beta_b}
  \min  
  \brx{
    \frac{q_b^2}{p_a} 
    ,\,
    \frac{1 - q_b}{\gamma p_a}
  }
  G(a, \Phi_{ax})_b^2
  \nn
  &
  \leq 
  2
  \sumak
  \sum_{b=1}^k
  \frac{1}{\beta_b}
  \frac{1}{u_a}
  \min  
  \brx{
    q_b^2
    ,\,
    \frac{1 - q_b}{\gamma}
  }
  G(a, \Phi_{ax})_b^2
  \tag{by $p_a \geq u_a / 2$}
  \nn
  &=
  \sumak
  \sum_{b=1}^k
  \frac{1}{\beta_b}
  \frac{1}{u_a}
  \min  
  \brx{
    q_b^2
    ,\,
    \frac{1 - q_b}{\gamma}
  }
  \prx{
    \sum_{e \in \mathrm{path}_\mathscr{T}(b)} w_e(a, \Phi_{ax})
  }^2
  \nn
  &\leq 
  \frac{m^2}{2}  
  \sumak
  \sum_{b=1}^k
  \frac{1}{\beta_b}
  \frac{1}{u_a}
  \min  
  \brx{
    q_b^2
    ,\,
    \frac{1 - q_b}{\gamma}
  }
  \prx{
    \sum_{e \in \mathrm{path}_\mathscr{T}(b)} \ind{a \in \calN_e}
  }^2
  \tag{by $\nrm{w_e}_\infty \leq m /2$}
  \nn 
  &\leq 
  2 m^2
  \sumak
  \sum_{b=1}^k
  \frac{1}{\beta_b}
  \frac{q_b}{u_a}
  \min  
  \brx{
    q_b
    ,\,
    \frac{1 - q_b}{\gamma q_b}
  }
  \ind{ a \in \cup_{e \in \mathrm{path}_\mathscr{T}(b)} \calN_e }
  \tag{since any action is in $\calN_e$ at most two edges in $e \in \mathrm{path}_\mathscr{T}(b)$}
  \nn
  &\leq 
  2 m^2
  \sumak
  \sum_{b=1}^k
  \frac{1}{\beta_b}
  \frac{q_b}{u_b}
  \min  
  \brx{
    q_b
    ,\,
    \frac{1 - q_b}{\gamma q_b}
  }
  \ind{ a \in \cup_{e \in \mathrm{path}_\mathscr{T}(b)} \calN_e }
  \tag{by $u_a \geq u_b$ for $a \in \mathrm{path}_\mathscr{T}(b)$ since Part (c) of Lemma~\ref{lem:wto}}
  \nn 
  &\leq 
  2 m^2 k
  \sumak
  \sum_{b=1}^k
  \frac{1}{\beta_b}
  \min  
  \brx{
    q_b
    ,\,
    \frac{1 - q_b}{\gamma q_b}
  }
  \ind{ a \in \cup_{e \in \mathrm{path}_\mathscr{T}(b)} \calN_e }
  \tag{by Part (b) of Lemma~\ref{lem:wto}}
  \nn 
  &=
  2 m^2 k^2
  \sum_{b=1}^k
  \frac{1}{\beta_b}
  \min  
  \brx{
    q_b
    ,\,
    \frac{1 - q_b}{\gamma q_b}
  }  
  \per
  \n
\end{align}

\paragraph{Step 5. Bounding the transformation term}
Owing to the property of the water transfer operator,
\begin{align}
  (p - q)^\top \lossmat \nu   
  =
  (1 - \lambda) (u - q)^\top \lossmat \nu   
  \leq 
  0
  \com 
  \n
\end{align}
where the last inequality follows by Part (a) of Lemma~\ref{lem:wto}.

Summing up the arguments for bounding the stability and transformation terms completes the proof of Lemma~\ref{lem:stab_bound_pm_lb}.
\end{proof}

\section{Deferred Proofs for Locally Observable Games (Section~\ref{sec:local})}
This section provides the deferred proofs for locally observable games.
Recall that the learning rate in~\eqref{eq:def_lr_local} is defined as 
\begin{equation}
  \alpha_{ta} = \min\brx{q_{ta}, \frac{1-q_{ta}}{\gamma q_{ta}}} \com
  \,\quad 
  \beta'_{ta} = c \sqrt{ \alpha_0 + \frac{1}{\gamma} \sum_{s=1}^{t-1} \alpha_{sa}}
  \com 
  \, \quad 
  \beta_{ta} = \max\{4 m k , \beta'_{ta}\}
  \com 
  \n
\end{equation}
where $\alpha_0 = \frac{1}{\gamma^{3/2}} + \epsilon$. 

\subsection{Proof of Lemma~\ref{lem:lob_1}}

\begin{proof}[Proof of Lemma~\ref{lem:lob_1}]
Recall that $a^* = \argmin_{a \in [k]} \E \brk[\big]{\sumT \lossmat_{a x_t}} \in \Pi$ is the optimal action in hindsight.
Then the regret is decomposed as
\begin{align}\label{eq:lob_1_1}
  \Reg_T
  &= 
  \Expect{\sumT (\lossmat_{A_t x_t} - \lossmat_{a^* x_t})}  
  = 
  \Expect{\sumT \sum_{b=1}^k p_{tb} (\lossmat_{b x_t} - \lossmat_{a^* x_t})}  \nn
  &= 
  \Expect{
  \sumT \sum_{b=1}^k (p_{tb} - q_{tb}) (\lossmat_{b x_t} - \lossmat_{a^* x_t})
  +
  \sumT \sum_{b=1}^k q_{tb} (\lossmat_{b x_t} - \lossmat_{a^* x_t})
  }
  \nn 
  &=
  \Expect{
  \sumT \sum_{b=1}^k (p_{tb} - q_{tb}) \lossmat_{b x_t}
  +
  \sumT \sum_{b=1}^k q_{tb} (\hat{y}_{tb} - \hat{y}_{ta^*})
  }
  \per
\end{align}

The second term of the last equality is bounded from above as
\begin{align}
  \Expect{
    \sumT \sum_{b=1}^k q_{tb} (\hat{y}_{tb} - \hat{y}_{ta^*})
  }
  =
  \Expect{
    \sumT \innerprod{q_t - e_{a^*}}{\hat y_t}
  }
  &=
  \Expect{
    \sumT \innerprod{q_t - u}{\hat y_t}
    +
    \sumT \innerprod{u - e_{a^*}}{\hat y_t}
  }
  \nn
  &\leq 
  \Expect{
    \sumT \innerprod{q_t - u}{\hat y_t}
  }
  +
  k
  \com 
  \label{eq:lob_1_3}
\end{align}
where the last inequality follows since
\begin{align}
  &
  \Expect{
    \innerprod{u - e_{a^*}}{\hat y_t}
  }
  =
  \Expect{
    \innerprod{u - e_{a^*}}{\sum_{c=1}^k G_t(c, \Phi_{c x_t})}
  }
  \nn 
  &=
  \Expect{
    \innerprod{u - e_{a^*}}{ \calL e_{x_t} + \delta \ones}
  }
  =
  \Expect{
    \frac{k}{T} 
    \innerprod{e_{a^*} - {\ones}/{k}} 
    { \calL e_{x_t} }
  }
  \leq 
  \frac{k}{T}
  \com 
  \label{eq:lob_1_4}
\end{align}
where the second equality holds since by Lemma~\ref{lem:Gdiff_Ldiff} there exists $\delta \in \R$ such that $\sum_{c=1}^k G_t(c, \Phi_{c x_t}) = \calL e_{x_t} + \delta \ones$.
Combining \eqref{eq:lob_1_1} and \eqref{eq:lob_1_3} with Lemmas~\ref{lem:ftrl} and~\ref{lem:stab_bound_pm_lb} completes the proof.
\end{proof}

\subsection{Proof of Lemma~\ref{lem:lob_pen_tr_stab}}
\begin{proof}[Proof of Lemma~\ref{lem:lob_pen_tr_stab}]

First we consider bounding the sum of the optimal value $\OPT_{q_t}(\beta_t)$.
The analysis was done in Section~\ref{subsec:exo_hybrid}; By Lemma~\ref{lem:stab_bound_pm_lb}, the first summation is bounded by
\begin{align}  
  \max\{ 0,\, \OPT_{q_t}(\beta_t) \}
  \leq 
  2 m^2 k^2 \sumak \frac{\alpha_{ta}}{\beta_{ta}}
  \leq 
  2 m^2 k^2 \sumak \frac{\alpha_{ta}}{\beta'_{ta}}
  \per 
  \n
\end{align}  

Next we consider the penalty term.
It holds that $\psi_t(q_{t+1}) - \psi_{t+1}(q_{t+1}) \leq 0$ since $\psi_t(p) \leq \psi_{t+1}(p)$ for all $p \in \calP_k$.
Since $\psi_t \geq 0$ it suffices to bound $\psi_{T+1}(u)$.
By the definition of the regularizer $\psi_t$ in~\eqref{eq:defpsilocal}, we have
\begin{align}
  \psi_{t} (u)
  =
  \sumak \beta_{ta} \phi(u_a)
  \leq
  \max_{x \in [1/T, 1]}  \phi(x) \sumak \beta_{ta} 
  \leq
  \max \{ \phi(1/T), \phi(1) \}
  \sumak \beta_{ta}
  \com
  \n
\end{align}
where the first inequality follows by $u_i \geq 1/T$ for all $i \in [k]$ and the second inequality holds since $\phi$ is convex.
Further, from the definition of $\phi(x) = (\phiLB(x) + x - 1) + \gamma \cdot (\phiCNS(x) + x) $ in~\eqref{eq:defpsilocal},
\begin{align*}
  \max \{ \phi(1/T), \phi(1) \}
  &
  =
  \max \left\{
     \log T + \frac{1}{T} - 1 + \gamma \left[ \left( 1 - \frac{1}{T} \right) \log \left( 1 - \frac{1}{T} \right) + \frac{1}{T} \right], \,
    \gamma
  \right\}
  \\
  &
  \leq
  \max \left\{
    \log T 
    +
    \frac{1+\gamma}{T} - 1 ,\,
    \gamma
  \right\}
  = \gamma
  \per 
\end{align*}
Hence, $\psi_{T+1} (u) \leq \gamma \sumak \beta_{T+1,a}$.
Summing up the arguments so far, we can bound the penalty term by
\begin{align}
  \psi_{T+1}(u)
  \leq 
  \gamma \sumak \beta_{T+1, a} 
  \leq 
  \gamma \sumak \beta'_{T+1, a} 
  +
  4 m k^2 \gamma 
  \com 
  \n
\end{align}
which follows by the definition of $\beta_{ta}$ and completes the proof of Lemma~\ref{lem:lob_pen_tr_stab}.
\end{proof}

\subsection{Proof of Theorem~\ref{thm:lob-main-theorem}}\label{subsec:proof-lob-main-thm}
Finally, we are ready to prove Theorem~\ref{thm:lob-main-theorem} with Lemmas~\ref{lem:lob_1} and~\ref{lem:lob_pen_tr_stab}.
\begin{proof}[Proof of Theorem~\ref{thm:lob-main-theorem}]
We first bound the RHS of $\OPT_{q_t}(\beta_t) \leq 2 m^2 k^2 \sumak \frac{\alpha_{ta}}{\beta'_{ta}}$ in Lemmas~\ref{lem:lob_pen_tr_stab} 
from the definition of $\beta_t$ and $\alpha_t$ in the following.
Since $\alpha_0 \geq \frac{1}{\gamma^{3/2}} \geq \frac{1}{\gamma} \alpha_t$,
\begin{align}
  \sumT \frac{\alpha_{ta}}{\beta'_{ta}} 
  &=
  \frac{1}{c} \sumT \frac{\alpha_{ta}}{\sqrt{\alpha_0 + \frac{1}{\gamma} \sum_{s=1}^{t-1} \alpha_{ta}}}
  \leq 
  \frac{\gamma}{c} \sumT \frac{\alpha_{ta} / \gamma }{\sqrt{\alpha_0 - \frac{1}{\gamma^{3/2}} + \frac{1}{\gamma} \sum_{s=1}^{t} \alpha_{ta}}}
  \nn 
  &\leq 
  \frac{2\gamma }{c} \prx{
    \sqrt{ \alpha_0 - \frac{1}{\gamma^{3/2}} + \frac1\gamma \sumT \alpha_{ta} }
    -
    \sqrt{ \alpha_0 - \frac{1}{\gamma^{3/2}} }
  }
  \leq
  \frac{2\gamma}{c^2} \prx{ \beta'_{T+1, a} - \beta'_{1,a} }
  +
  \frac{2}{c \gamma^{5/4}}
  \com 
  \label{eq:lob_thm_p_1}
\end{align}
where the last inequality follows since
\begin{align}
  \frac{2}{c^2} \beta'_{1,a} - \frac{2}{c} \sqrt{\alpha_0 - \frac{1}{\gamma^{3/2}}}
  =
  \frac{2}{c} \prx{ \sqrt{\alpha_0} - \sqrt{\alpha_0 - \frac{1}{\gamma^{3/2}}}}
  \leq 
  \frac{2}{c} \frac{1}{\gamma^{3/2}} \frac{1}{ \sqrt{\alpha_0} + \sqrt{\alpha_0 - \frac{1}{\gamma^{3/2}}}}
  \leq 
  \frac{2}{c \gamma^{9/4}}
  \per 
  \n
\end{align}
Hence combining Lemmas~\ref{lem:lob_1} and~\ref{lem:lob_pen_tr_stab} with~\eqref{eq:lob_thm_p_1} gives
\begin{align}
  \Reg_T
  &\leq
  \Expect{
    \sumT \gamma \prx{1 + \frac{4 m^2 k^2}{c^2}} \sumak 
    \beta'_{T+1, a}
  }
  +
  4 m k^2 \gamma 
  +
  \frac{4 m^2 k^2}{c \gamma^{5/4}}
  +
  \frac{mk^3}{2}
  \nn
  &=
  \Expect{
    2 \gamma
    \sumT \sumak 
    \beta'_{T+1, a}
  }
  +
  4 m k^2 \gamma 
  +
  \frac{4 m^2 k^2}{c \gamma^{5/4}}
  \com 
  \label{eq:lob_thm_p_2}
\end{align}
where the equality follows from $c = 2 mk$.

In the following, we bound $\beta'_{T+1,a}$ for suboptimal arm $a \neq a^*$ and optimal arms $a^*$ separately.
For suboptimal arm $a \neq a^*$, the definition of $\beta'_{ta}$ in~\eqref{eq:def_lr_local} implies that
\begin{align}
  \beta'_{T+1,a}
  \leq 
  c \sqrt{ \alpha_0  + \frac{\sum_{t=1}^{T} q_{ta} }{\gamma}}
  \per 
  \label{eq:beta_bound_subopt}
\end{align}
Next we consider optimal action $a^*$.
Since $\sqrt{\gamma} \geq 2$ for any $z \in [0,1]$, it holds that
\begin{align}
  \min  
  \brx{
    z
    ,\,
    \frac{1 - z}{\gamma z}
  }
  \leq 
  \begin{cases}
    z & \mbox{if} \; z \leq \frac{1}{\sqrt{\gamma}}  \\
    \frac{1 - z}{\sqrt{\gamma}}  &  \mbox{otherwise}
  \end{cases}
  \leq 
  \frac{2}{\sqrt{\gamma}} (1 - z)
  \per 
 \n 
\end{align}
Using this inequality, 
\begin{align} 
  \sumT \alpha_{ta^*}  
  \leq
  \frac{2}{\sqrt{\gamma}} \sumT (1 - q_{ta^*})
  =
  \frac{2}{\sqrt{\gamma}} \sumT \sum_{a \neq a^*} q_{ta}
  \com 
 \n 
\end{align}
which implies
\begin{align}
  \beta'_{T+1, a^*}
  \leq
  c \sqrt{ \alpha_0 + \frac{2}{\gamma^{3/2}} \sumT \sum_{a \neq a^*} q_{ta} }
  \per 
  \label{eq:beta_bound_opt}
\end{align}

Summing up the arguments so far by using~\eqref{eq:beta_bound_subopt} and~\eqref{eq:beta_bound_opt} with~\eqref{eq:lob_thm_p_2}
and letting $Q_a = \Expect{\sumT q_{ta}}$,
we can bound the regret as 
\begin{align}
  \Reg_T
  &\leq   
  2 \gamma 
  \sum_{a \neq a^*}
  \Expect{
    c \sqrt{\alpha_0 +  \sumT \frac{q_{ta}}{\gamma}}
  }
  +
  4 m k^2 \gamma
  +
  \gamma c \sqrt{ \alpha_0 + \frac{2}{\gamma^{3/2}} \sumT \sum_{a \neq a^*} q_{ta} }
  +
  \frac{4 m^2 k^2}{c \gamma^{5/4}}
  \nn
  &\leq   
  2 \gamma 
  \sum_{a \neq a^*}
  c \sqrt{\alpha_0 +  \frac{1}{\gamma} \Expect{\sumT q_{ta}}}
  +
  4 m k^2 \gamma
  +
  \gamma c \sqrt{ \alpha_0 + \frac{2}{\gamma^{3/2}} \sum_{a \neq a^*} \Expect{ \sumT q_{ta} }  }
  +
  \frac{4 m^2 k^2}{c \gamma^{5/4}}
  \nn
  &\leq   
  2 \gamma 
  \sum_{a \neq a^*}
  c \sqrt{\alpha_0 +  \frac{Q_a}{\gamma} }
  +
  4 m k^2 \gamma
  +
  \gamma c \sqrt{ \alpha_0 + \frac{2}{\gamma^{3/2}} \sum_{a \neq a^*} Q_a }
  +
  \frac{4 m^2 k^2}{c \gamma^{5/4}}
  \per 
  \label{eq:lob_p_final}
\end{align}

For the adversarial environment, using the Cauchy--Schwarz inequality and $\sumak q_{ta} = 1$ gives
\begin{align}
  \Reg_T
  &\leq 
  2 \gamma c \sqrt{ k \prx{\alpha_0 + \frac{T}{\gamma} } }
  + 
  4 m k^2 \gamma 
  +
  \gamma c \sqrt{ k \prx{  \alpha_0 + \frac{2 T }{\gamma^{3/2}}} }
  +
  \frac{4 m^2 k^2}{c \gamma^{5/4}}
  \nn
  &
  \leq 
  6 m k^{3/2} \sqrt{T \log T }
  +
  4 m k^2 \log T 
  +
  12 mk \sqrt{k \alpha_0 \log T}
  +
  \frac{2 m k}{(\log T)^{5/4}}
  \per 
  \n 
\end{align}

For the stochastic environment with adversarial corruptions, we have $\Reg_T \geq \frac{1}{2} \sum_{a \neq a^*} \Delta_a Q_a - C$ by $p_t \in \calR(q)$.
Hence for any $\lambda \in (0,1]$, we have
\begin{align}
  \Reg_T 
  &= 
  (1 + \lambda) \Reg_T - \lambda \Reg_T
  \nn
  &\leq 
  2 (1 + \lambda) \gamma 
  \sum_{a \neq a^*}
  c \sqrt{\alpha_0 +  \frac{Q_a}{\gamma} }
  +
  (1 + \lambda) \gamma c \sqrt{ \alpha_0 + \frac{2}{\gamma^{3/2}} \sum_{a \neq a^*} Q_a }
  - \frac{\lambda}{2} \sum_{a \neq a^*} \Delta_a Q_a
  \nn 
  &\qquad 
  +
  (1 + \lambda) m k^2 \gamma
  +
  O(mk)
  +
  \lambda C
  \tag{By \eqref{eq:lob_p_final} and $\Reg_T \geq \frac{1}{2} \sum_{a \neq a^*} \Delta_a Q_a$}
  \nn 
  &\leq
  \sum_{a \neq a^*}
  \prx{
    2 (1 + \lambda) c \sqrt{ \gamma Q_a }
    -
    \frac{\lambda}{4} \Delta_a Q_a
  }
  +
  (1 + \lambda)  c \sqrt{ {2 \sqrt{\gamma}} \sum_{a \neq a^*} Q_a }
  - \frac{\lambda}{4} \Deltamin \sum_{a \neq a^*} Q_a
  \nn 
  &\qquad 
  +
  (1 + \lambda) m k^2 \gamma
  +
  12 mk \sqrt{k \alpha_0 \log T}
  +
  O(mk)
  +
  \lambda C
  \nn 
  &\leq 
  \sum_{a \neq a^*}
  \frac{64 (1 + \lambda)^2 c^2}{\lambda} \frac{\gamma}{\Delta_a}
  +
  \frac{32 (1+\lambda)^2 c^2}{\lambda} \frac{\sqrt{\gamma}}{\Deltamin}
  +
  (1 + \lambda) m k^2 \gamma
  +
  12 mk \sqrt{k \alpha_0 \log T}
  +
  O(mk)
  + 
  \lambda C
  \com 
 \n 
\end{align}
where the last inequality follows from $bx - ax^2 \leq b^2 / (4a)$ for $a > 0$ and $b \geq 0$.
Setting optimal $\lambda$ completes the proof of Theorem~\ref{thm:lob-main-theorem}.
\end{proof}

\section{Deferred Proofs for Globally Observable Games (Theorem~\ref{thm:gob-main-theorem}, Section~\ref{sec:global})}\label{sec:proof-global}

\begin{algorithm}[t]
\caption{
Algorithm for globally observable games
}
\label{alg:global}
\begin{algorithmic}    
\FOR{$t = 1, 2, \ldots$} 
  \STATE Compute $\qt$ via FTRL~\eqref{eq:def_q} with regularizer~\eqref{eq:defpsiglobal}.
  \STATE Compute $\pt$ from $\qt$ by~\eqref{eq:p_global}. 
  \STATE Sample $\at \sim \pt$, observe $\Phi_{A_t x_t} \in \Sigma$, and compute $\displaystyle \hat{y}_t = {G^\circ(A_t, \sigma_t)}/{\ptAt}$.
\ENDFOR
\end{algorithmic}
\end{algorithm}

This appendix provides the proof of Theorem~\ref{thm:gob-main-theorem}.
Let $u \in \calP_k$ be the projection of $e_{a^*}$ on $[1/T, 1 - 1/T]^k \cap \calP_k$. In particular, define $u \in [1/T, 1 - 1/T]^k \cap \calP_k$ by 
\begin{equation}
  u = \argmin_{u' \in [1/T, 1 - 1/T]^k \cap \calP_k} \nrm{u' - e_{a^*}}_{\infty}
  \per 
 \n 
\end{equation}
We first decompose the regret as follows.
\begin{lemma}\label{lem:gob_1}
The regret of Algorithm~\ref{alg:global} is bounded as 
$
  \Reg_T
  \leq 
  \E\big[
    \sumT
    \gamma_t
  +
    \sumT
    \prn[\big]{
    \innerprod{\hat{y}_t}{\qt - q_{t+1}}
    -
    D_{\psi_t}(q_{t+1}, \qt)
    }
  +
    \sumT
    \prn[\big]{
    \psi_{t}(q_{t+1})
    -
    \psi_{t+1}(q_{t+1})
    }
    +
    \psi_{T+1} (u) 
    -
    \psi_1 (q_1)
  \big]
  +
  c_{\calG}
  \per
$
\end{lemma}
This lemma can be proven in a very similar manner as Lemma~\ref{lem:lob_1}.
Note that the first, second, and third summations correspond to the transformation, stability, and penalty terms, respectively.

\begin{proof}[Proof of Lemma~\ref{lem:gob_1}]
We first decompose the regret as follows:
\begin{align}
  \Reg_T
  &= 
  \Expect{\sumT \prx{\lossmat_{\at x_t} - \lossmat_{a^* x_t}}}
  =
  \Expect{\sumT \innerprod{\pt - e_{a^*}}{\lossmat e_{x_t}}}
  =
  \Expect{
  \sumT \innerprod{\qt - e_{a^*}}{\lossmat e_{x_t}}
  + 
  \sumT \gamma_t \innerprod{\frac1k \ones - \qt}{\lossmat e_{x_t}}
  }
  \nn
  &
  \le
  \Expect{
  \sumT \innerprod{\qt - e_{a^*}}{\lossmat e_{x_t}}
  + 
  \sumT \gamma_t 
  }
  =
  \Expect{
  \sumT \sumak \qta \prx{\lossmat_{a x_t} - \lossmat_{a^* x_t}}
  +
  \sumT \gamma_t 
  }
  \nn
  &
  =
  \Expect{
  \sumT \sumak \qta \prx{\hat y_{t a} - \hat y_{t a^*}}
  +
  \sumT \gamma_t 
  }
  =
  \Expect{
  \sumT \innerprod{q_t - e_{a^*}}{\hat y_t}
  +
  \sumT \gamma_t 
  }
  \com 
  \label{eq:gob_1_1}
\end{align}
where the inequality follows from the fact that $\lossmat \in [0,1]^{k \times d}$, 
and the fifth equality follows from the definitions of $\hat{y}_t$ and $\qta = 0$ for $a \not\in \Pi$.
The first term in~\eqref{eq:gob_1_1} is further bounded as
\begin{align}
  \Expect{
    \sumT \innerprod{q_t - e_{a^*}}{\hat y_t}
  }
  =
  \Expect{
    \sumT \innerprod{q_t - u}{\hat y_t}
    +
    \sumT \innerprod{u - e_{a^*}}{\hat y_t}
  }
  \leq 
  \Expect{
    \sumT \innerprod{q_t - u}{\hat y_t}
  }
  +
  k
  \com  
  \label{eq:gob_1_2}
\end{align}
where the last inequality follows by the same argument in~\eqref{eq:lob_1_4}.
Combining \eqref{eq:gob_1_1} and \eqref{eq:gob_1_2} with Lemma~\ref{lem:ftrl} completes the proof of Lemma~\ref{lem:gob_1}.
\end{proof}

We can bound the penalty and stability terms on the RHS of Lemma~\ref{lem:gob_1} as follows.
\begin{lemma}\label{lem:gob_pen_stab}
The penalty term is bounded by
$
  \sumT (\psi_t(q_{t+1}) - \psi_{t+1}(q_{t+1}))
  +
  \psi_{T+1}(u)
  -
  \psi_1(q_1)
  \leq 
  2 \beta_{T+1} k \log T
  \per 
$
The stability term is bounded by
\begin{align}
  \innerprod{q_t - q_{t+1}}{\hat{y}_t} - D_{\psi_t}(q_{t+1}, q_t)
  &\leq
  \frac{1}{\beta_t}
  \sum_{b=1}^k
  \min\brx{ q_{tb}^2, (1 - q_{tb})^2 }  \hat{y}_{tb}^2
  \per 
 \n 
\end{align}
\end{lemma}

\begin{proof}[Proof of Lemma~\ref{lem:gob_pen_stab}]
The penalty term is bounded by
\begin{align}
  \sumT (\psi_t(q_{t+1}) - \psi_{t+1}(q_{t+1}))
  +
  \psi_{T+1}(u)
  -
  \psi_1(q_1)
  \leq
  \psi_{T+1}(u)
  \leq
  2 \beta_{T+1} k \log T
  \com
 \n 
\end{align}
where the last inequality holds since $u \in [1/T, 1-1/T]^k$.
Next we consider the stability term.
Taking the worst case with respect to $q_{t+1}$,
\begin{align}
  &
  \innerprod{q_t - q_{t+1}}{\hat{y}_t} - D_{\psi_t}(q_{t+1}, q_t)
  =
  \sum_{b=1}^k
  \prx{
  \prx{q_{tb} - q_{t+1,b}} \hat{y}_{tb} 
  - 
  \beta_t \prx{ D_{\phiLB}(q_{t+1,b}, q_{tb} ) + D_{\phiCLB}(q_{t+1,b}, q_{tb} ) }
  }
  \nn
  &\leq 
  \sum_{b=1}^k
  \min \brx{
  \prx{
  \prx{q_{tb} - q_{t+1,b}} \hat{y}_{tb}
  - 
  \beta_t  D_{\phiLB}(q_{t+1,b}, q_{tb} ) 
  ,\,
  \prx{q_{tb} - q_{t+1,b}} \hat{y}_{tb}
  - 
  \beta_t  D_{\phiCLB}(q_{t+1,b}, q_{tb} ) 
  }
  }
  \nn 
  &\leq
  \frac{1}{\beta_t}
  \sum_{b=1}^k
  \min\brx{ q_{tb}^2, (1 - q_{tb})^2 }  \hat{y}_{tb}^2
  \per 
 \n 
\end{align}
The last inequality can be proven from Propositions~\ref{prop:stab_lb} and~\ref{prop:stab_clb} as follows. 
For fixed $b \in [k]$ and any $q_{tb} \leq 1/2$, we have
\begin{align}
  \prx{q_{tb} - q_{t+1,b}} \hat{y}_{tb}
  - 
  \beta_t  D_{\phiLB}(q_{t+1,b}, q_{tb} ) 
  &=
  \beta_t 
  \prx{
  \prx{q_{tb} - q_{t+1,b}} \frac{\hat{y}_{tb}}{\beta_t}
  - 
  D_{\phiLB}(q_{t+1,b}, q_{tb} ) 
  }
  \nn
  &\leq 
  \frac{1}{\beta_t} q_{tb}^2 \hat{y}_{tb}^2 
  \leq 
  \frac{1}{\beta_t} \min\brx{ q_{tb}^2, (1 - q_{tb})^2 } \hat{y}_{tb}^2 
  \com 
 \n 
\end{align}
where 
the second inequality holds by $q_{tb} \leq 1/2$ and
the first inequality holds from Proposition~\ref{prop:stab_lb} with
\begin{align}
  \abx{ \frac{\hat{y}_{tb}}{\beta_t} } 
  \leq 
  \frac{\nrm{G^\circ}_{\infty}}{\abx{\beta_t \, p_{tA_t} }}
  \leq 
  \nrm{G^\circ}_{\infty}
  \abx{\frac{k}{\beta_t \gamma_t }}
  \leq 
  \nrm{G^\circ}_{\infty}
  \abx{\frac{k}{ \sqrt{2 \beta_t} c_{\calG} z_t }}
  \leq 
  \nrm{G^\circ}_{\infty}
  \abx{\frac{k}{ \sqrt{2 \beta_t} c_{\calG} q_{tb} }}
  \leq
  \frac{1}{2 q_{tb}}
  \per 
 \n 
\end{align}
Here the forth inequality holds due to $z_t \geq \min \brx{{q_{tb}, 1-q_{tb}}} = q_{tb}$ and the last inequality follows by $\beta_t \geq \frac{2k^2}{c_{\calG}^2}$.
For fixed $b \in [k]$ and any $q_{tb} > 1/2$ it also holds that
\begin{align}
  \prx{q_{tb} - q_{t+1,b}} \hat{y}_{tb}
  - 
  \beta_t  D_{\phiCLB}(q_{t+1,b}, q_{tb} ) 
  \leq 
  \frac{1}{\beta_t} (1-q_{tb})^2 \hat{y}_{tb}^2 
  \leq 
  \frac{1}{\beta_t} \min\brx{ q_{tb}^2, (1 - q_{tb})^2 } \hat{y}_{tb}^2 
  \com 
 \n 
\end{align}
where the last inequality follows by $q_{tb} > 1/2$, and
the first inequality holds from
Propositions~\ref{prop:stab_lb} with
\begin{align}
  \abx{ \frac{\hat{y}_{tb}}{\beta_t} } 
  \leq 
  \frac{\nrm{G^\circ}_{\infty}}{\abx{\beta_t \, p_{tA_t} }}
  \leq 
  \nrm{G^\circ}_{\infty}
  \abx{\frac{k}{\beta_t \gamma_t }}
  \leq 
  \nrm{G^\circ}_{\infty}
  \abx{\frac{k}{ \sqrt{2 \beta_t} c_{\calG} z_t }}
  \leq 
  \nrm{G^\circ}_{\infty}
  \abx{\frac{k}{ \sqrt{2 \beta_t} c_{\calG} (1 - q_{tb}) }}
  \leq
  \frac{1}{2 (1 - q_{tb})}
  \per  
 \n 
\end{align}
\end{proof}

Finally we are ready to prove Theorem~\ref{thm:gob-main-theorem}.
\begin{proof}[Proof of Theorem~\ref{thm:gob-main-theorem}]
First, taking expectation with respect to $A_t \sim p_t$,
we can further bound the stability term as
\begin{align}
  \Et{
    \frac{1}{\beta_t}
    \sum_{b=1}^k
    \min\brx{ q_{tb}^2, (1 - q_{tb})^2 }  \hat{y}_{tb}^2
  }  
  =
  \frac{1}{\beta_t}
  \sumak
  \frac{1}{p_{ta}}
  \sum_{b=1}^k
  \min\brx{ q_{tb}^2, (1 - q_{tb})^2 } G(a, \Phi_{a x_t})_b^2
  \leq 
  \frac{c_{\calG}^2}{\beta_t \gamma_t} z_t^2
  \per
 \n 
\end{align}
Combining this inequality with Lemmas~\ref{lem:gob_1} and \ref{lem:gob_pen_stab} yields
\begin{align}
  \Reg_T
  &\leq 
  \Expect{
    2 \beta_{T+1} k \log T
    +
    \sumT \prx{
      \frac{c_{\calG}^2}{\gamma_t \beta_t} z_t^2
      +
      \gamma_t
    }
  }
  + c_{\calG}
  =
  \Expect{
    2 \beta_{T+1} k \log T
    +
    2 c_{\calG}  \sumT \frac{ z_t}{\sqrt{\beta_t}}
  }
  + k
  \per 
 \n 
\end{align}
From the definition of $\beta_t$,
\begin{align}
  \sqrt{c_1}
  \sumT \frac{z_t}{\sqrt{\beta_t}}
  =
  \sumT \frac{z_t}{ \prx{k + \sum_{s=1}^{t-1} z_s}^{1/3} }
  \leq 
  \sumT \frac{z_t}{ \prx{\sum_{s=1}^{t} z_s}^{1/3} }
  \leq 
  \int_{0}^{\sumT z_t} x^{-1/3} \d x
  =
  \frac{3}{2} \prx{ \sumT z_t }^{2/3}
  \per 
 \n
\end{align}
Using this inequality and Jensen's inequality,
\begin{align}
  \Reg_T
  \leq
  \prx{ 2 c_1 k \log T + \frac{3 c_{\calG}}{\sqrt{c_1}} }
  \prx{k + \Expect{ \sumT z_t} }^{2/3}
  =
  O \prx{
  c_{\calG}^{2/3} \prx{k \log T}^{1/3}
  \prx{k + \Expect{ \sumT z_t } }^{2/3}
  }
  \per 
 \n
\end{align}

For adversarial environments, since $z_t \leq 1$, $\Reg_T = O\prn[\big]{ \prn{c_{\calG}^2 k \log T}^{1/3} T^{2/3} }$.
For stochastic environments, using $z_t \leq \min_{b\in[k]} \sum_{a \neq b} q_{ta} \leq 1 - q_{ta^*}$, for any $\lambda \in (0,1]$,
\begin{align}
  \Reg_T
  &=
  (1 + \lambda) \Reg_T - \lambda \Reg_T
  \nn 
  &\leq 
  O\prx{
  (1 + \lambda)
  c_{\calG}^{2/3} \prx{k \log T}^{1/3}
  \prx{k + \Expect{ \sumT (1 - q_{ta^*}) } }^{2/3}
  }
  - 
  \frac{\lambda}{2} \Delta \, \Expect{\sumT (1 - q_{ta^*})}
  +
  \lambda C
  \nn 
  &\leq
  O\prx{ 
  \frac{(1 + \lambda)^3}{\lambda^2}
  \frac{c_{\calG}^2 k \log T}{\Delta^2}
  +
  c_{\calG}^{2/3} \prx{\log T}^{1/3} k^{4/3}
  }
  +
  \lambda C
  \nn 
  &\leq
  O\prx{
  \frac{c_{\calG}^2 k \log T}{\Delta^2}
  +
  \frac{1}{\lambda^2}
  \frac{c_{\calG}^2 k \log T}{\Delta^2}
  +
  c_{\calG}^{2/3} \prx{\log T}^{1/3} k^{4/3}
  }
  +
  \lambda C
  \com 
 \n
\end{align}
where the second inequality follows by $b x^2 - a x^3 \leq \frac{4 b^3}{27 a^2}$ and the last inequality follows by $\lambda \in (0,1]$.
Setting $\lambda = O\prn[\Big]{ \prx{ \frac{c_{\calG}^2 k \log T}{C} }^{1/3} }$ completes the proof of Theorem~\ref{thm:gob-main-theorem}.
\end{proof}


\end{document}